\documentclass{article}

\PassOptionsToPackage{}{natbib}

\usepackage[final]{neurips_2024}

\usepackage[utf8]{inputenc} 
\usepackage{enumitem} 
\usepackage[T1]{fontenc}    
\usepackage{hyperref}       
\usepackage{url}            
\usepackage{booktabs}       
\usepackage{amsfonts}       
\usepackage{nicefrac}       
\usepackage{microtype}      
\usepackage{xcolor}         
\usepackage{amsmath}
\usepackage{graphicx}
\usepackage{wrapfig}
\usepackage{adjustbox}
\usepackage{array}
\usepackage{caption}
\usepackage{hyperref} 
\usepackage{makecell}
\usepackage{algorithm}
\usepackage{algpseudocode}
\usepackage{tcolorbox}
\usepackage{subcaption}
\usepackage{listings}

\title{Improving the Training of Rectified Flows}

\newtheorem{proposition}{Proposition}

\newcommand{\ab}{\mathbf{a}}
\newcommand{\bb}{\mathbf{b}}
\newcommand{\x}{\mathbf{x}}

\newcommand{\z}{\mathbf{z}}
\newcommand{\vb}{\mathbf{v}}
\newcommand{\ub}{\mathbf{u}}

\newcommand{\zerob}{\mathbf{0}}
\newcommand{\Ib}{\mathbf{I}}
\newcommand{\thetab}{\boldsymbol{\theta}}
\author{%
  Sangyun Lee \\
  Carnegie Mellon University\\
  \texttt{sangyunl@andrew.cmu.edu} \\
\And
  Zinan Lin \\
  Microsoft Research \\
  \texttt{zinanlin@microsoft.com} \\
\And
Giulia Fanti \\
  Carnegie Mellon University\\
  \texttt{gfanti@andrew.cmu.edu} \\
}

\begin{document}

\maketitle

\begin{abstract}
    Diffusion models have shown great promise for image and video generation, but sampling from state-of-the-art models requires expensive numerical integration of a generative ODE.
    One approach for tackling this problem is rectified flows, which iteratively learn smooth ODE paths that are less susceptible to truncation error.
    However, rectified flows still require a relatively large number of function evaluations (NFEs).
    In this work, we propose improved techniques for training rectified flows, allowing them to compete with \emph{knowledge distillation} methods even in the low NFE setting.
    Our main insight is that under realistic settings, a single iteration of the Reflow algorithm for training rectified flows is sufficient to learn nearly straight trajectories; hence, the current practice of using multiple Reflow iterations is unnecessary.
    We thus propose techniques to improve one-round training of rectified flows, including a U-shaped timestep distribution and LPIPS-Huber premetric.
    With these techniques, we improve the FID of the previous 2-rectified flow by up to 75\% in the 1 NFE setting on CIFAR-10.
    On ImageNet 64$\times$64, our improved rectified flow outperforms the state-of-the-art distillation methods
    such as consistency distillation and progressive distillation in both one-step and two-step settings and rivals the performance of improved consistency training (iCT) in FID.
    Code is available at \url{https://github.com/sangyun884/rfpp}.
\end{abstract}

\section{Introduction}

Diffusion models~\citep{sohl2015deep,ho2020denoising,song2019generative,song2020score} have shown great promise in image~\citep{ramesh2022hierarchical} and video~\citep{ho2022video} generation.
They generate data by simulating a stochastic denoising process where noise is gradually transformed into data. To sample efficiently from diffusion models, the denoising process is typically converted into a counterpart of Ordinary Differential Equations (ODEs)~\citep{song2020score} called probability flow ODEs (PF-ODEs).

Despite the success of diffusion models using PF-ODEs, drawing high-quality samples requires numerical integration of the PF-ODE with small step sizes, which is computationally expensive.
Today, two prominent classes of approaches for tackling this issue are: (1) knowledge distillation (e.g., consistency distillation~\citep{song2023consistency}, progressive distillation~\citep{salimans2022progressive}) and (2) simulation-free flow models (e.g., rectified flows~\citep{liu2022flow}, flow matching~\citep{lipman2022flow}).

In knowledge distllation-based methods~\citep{luhman2021knowledge,salimans2022progressive,song2023consistency,zheng2022fast} a student model is trained to directly predict the solution of the PF-ODE.
These models are currently state-of-the-art in the low number of function evaluations (NFEs) regime (e.g. 1-4).

Another promising direction is simulation-free flow models such as rectified flows~\citep{liu2022flow,liu2022rectified},
a generative model that learns a transport map between two distributions defined via neural ODEs. 
Diffusion models with PF-ODEs are a special case.
Rectified flows can learn smooth ODE trajectories that are less susceptible to truncation error, which allows for high-quality samples with fewer NFEs than diffusion models.
They have been shown to outperform diffusion models in the moderate to high NFE regime~\citep{lipman2022flow,liu2022flow,esser2024scaling}, but they still require a relatively large number of NFEs compared to distillation methods.

Compared to knowledge distillation methods~\citep{luhman2021knowledge,salimans2022progressive,song2023consistency,zheng2022fast} rectified flows have several advantages.
First, they can be generalized to map two arbitrary distributions to one another, while distillation methods are limited to a Gaussian noise distribution.
Also, as a neural ODE, rectified flows naturally support \textit{inversion} from data to noise, which has many applications including image editing~\citep{hertz2022prompt,kim2022diffusionclip,wallace2023edict,couairon2022diffedit,mokady2023null,su2022dual,hong2023exact} and watermarking~\citep{wen2023tree}.
Further, the likelihood of rectified flow models can be evaluated using the instantaneous change of variable formula~\citep{chen2018neural}, whereas this is not possible with knowledge distillation-based methods.
In addition, rectified flows can flexibly adjust the balance between the sample quality and computational cost by altering NFEs, whereas distillation methods either do not support multi-step sampling or do not necessarily perform better with more NFEs (e.g. > 4)~\citep{kim2023consistency}.

Given the qualitative advantages of rectified flows, a natural question is, \textbf{can rectified flows compete with distillation-based methods such as consistency models~\citep{song2023consistency} in the low NFE setting?}
Today, the state-of-the-art techniques for training rectified flows use the \textit{Reflow} algorithm to improve low NFE performance~\citep{liu2022flow,liu2023instaflow}.
Reflow is a recursive training algorithm where the rectified flow is trained on data-noise pairs generated by the generative ODE of the previous stage model.
In current implementations of Reflow, to obtain a reasonable one-step generative performance, Reflow should be applied at least twice, followed by an optional distillation stage to further boost  performance~\citep{liu2022flow,liu2023instaflow}.
Each training stage requires generating a large number of data-noise pairs and training the model until convergence, which is computationally expensive and leads to error accumulation across rounds.
Even with these efforts, the generative performance of rectified flow still lags behind the distillation methods such as consistency models~\citep{song2023consistency}.

We show that \textbf{rectified flows can indeed be competitive with the distillation methods in the low NFE setting by applying Reflow with our proposed training techniques}.
Our techniques are based on the observation  that under realistic settings, the linear interpolation trajectories of the pre-trained rectified flow rarely intersect with each other.
This provides several insights: 1) applying Reflow once is sufficient to obtain straight-line generative ODE in the optima,
2) the training loss of 2-rectified flow has zero lower bound, 
and 3) other loss functions than the squared $\ell_2$ distance can be used during training.
Based upon this finding,
we propose several training techniques to improve Reflow, including: (1) a U-shaped timestep distribution, (2) an LPIPS-huber premetric, which we find to be critical for the few-step generative performance.
After being initialized with pre-trained diffusion models such as EDM~\citep{karras2022elucidating}, our method only requires one training stage without additional Reflow or distillation stages, unlike previous works~\citep{liu2022flow,liu2023instaflow}.

Our evaluation shows that on several datasets (CIFAR-10~\citep{krizhevsky2009learning}, ImageNet 64$\times$64~\citep{deng2009imagenet}),
our improved rectified flow outperforms the state-of-the-art distillation methods
such as consistency distillation (CD)~\citep{song2023consistency} and progressive distillation (PD)~\citep{salimans2022progressive} in both one-step and two-step settings, and it rivals the performance of the improved consistency training (iCT)~\citep{song2023consistency}
in terms of the Frechet Inception Distance~\citep{heusel2017gans} (FID).
Our training techniques \textbf{reduce the FID of the previous 2-rectified flow~\citep{liu2022flow} by about $\mathbf{75\%}$} ($12.21 \to 3.07$) on CIFAR-10.
Ablations on three datasets 
show that the proposed techniques give a consistent and sizeable gain.
We also showcase the qualitative advantages of rectified flow such as few-step inversion, and its application to interpolation and image-to-image translation.

\section{Background}
\subsection{Rectified Flow}

\begin{figure}[t]
    \centering
    \includegraphics[width=0.8\linewidth]{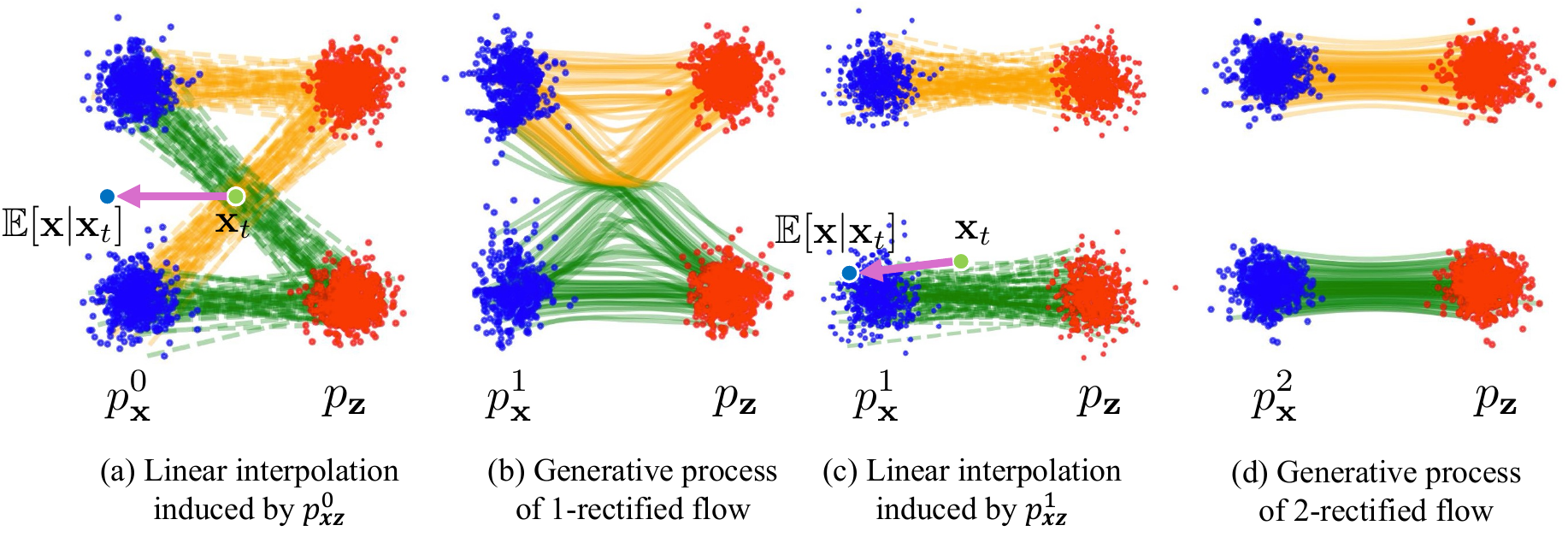}
    \vspace{-2mm}
    \caption{Rectified flow process (figure modified from \cite{liu2022flow}). Rectified flow rewires trajectories so there are no intersecting trajectories $(a)\to (b)$. Then, we take noise samples from $p_\z$ and their generated samples from $p^1_\x$, and linearly interpolate them $(c)$. In Reflow, rectified flow is applied again $(c)\to (d)$ to straighten flows. This procedure is repeated recursively.}
    \vspace{-2mm}
    \label{fig:rectified_flow}
\end{figure}

Rectified flow (see also flow matching~\citep{lipman2022flow} and stochastic interpolant~\citep{albergo2022building}) is a generative model that smoothly transitions between two distributions $p_{\x}$ and $p_{\z}$ by solving ODEs~\citep{liu2022flow}.
For $\x \sim p_{\x}$ and $\z \sim p_{\z}$, we define the interpolation between $\x$ and $\z$ as $\x_t = (1-t)\x + t\z$ for $t \in [0, 1]$.
\citet{liu2022flow} showed that for $\z_0 \sim p_\x$, the following ODE yields the same marginal distribution as $\x_t$ for any $t$:
\begin{align}
    \frac{d\z_t}{dt} = \vb_t(\z_t) &:= \frac{1}{t}(\z_t - \mathbb E[\x | \x_t = \z_t]).
    \label{eq:ode}
\end{align}
Since $\x_1 = \z$, Eq.~\eqref{eq:ode} transports $p_{\x}$ to $p_{\z}$.
We can also transport $p_{\z}$ to $p_{\x}$ by drawing $\z_1$ from $p_{\z}$ and solving the ODE backwards from $t=1$ to $t=0$.
During training, we estimate the conditional expectation $\mathbb E[\x | \x_t = \z_t]$ with a vector-valued neural network $\x_{\thetab}$ trained on the squared $\ell_2$ loss:
\begin{align}
    \min_{\thetab} \mathbb E_{\x, \z \sim p_{\x \z}} \mathbb E_{t \sim p_t} [\omega(t)||\x - \x_{\thetab} (\x_t, t)||_2^2],
    \label{eq:objective-x}
\end{align}
where $p_{\x \z}$ is the joint distribution of $\x$ and $\z$, $\x_{\thetab}$ is parameterized by $\thetab$, and $\omega(t)$ is a weighting function.
$p_t$ is chosen to be the uniform distribution on $[0, 1]$ in \citet{liu2022flow,liu2023instaflow}.
In the optimum of Eq.~\eqref{eq:objective-x}, $\x_{\thetab}$ becomes the conditional expectation as it is a minimum mean squared error (MMSE) estimator, which is then plugged into the ODE~\eqref{eq:ode} to generate samples.
Instead of predicting the conditional expectation directly, \citet{liu2022flow} choose to parameterize the velocity $\vb_t$ with a neural network $\vb_{\thetab}$ and train it on
\begin{align}
    \min_{\thetab} \mathbb E_{\x, \z \sim p_{\x \z}} \mathbb E_{t \sim p_t} [||(\z - \x) - \vb_{\thetab}(\x_t, t)||_2^2],
    \label{eq:objective-v}
\end{align}
which is equivalent to Eq.~\eqref{eq:objective-x} with $\omega(t) = \frac{1}{t^2}$. See Appendix.~\ref{sec:equivalence}.

In this paper, we consider the Gaussian marginal case, i.e., $p_\z = \mathcal N(\zerob, \Ib)$.
In this case, if we define $\x$ and $\z$ as independent random variables (i.e., $p_{\x \z}(\x,\z) = p_{\x}(\x) p_{\z} (\z)$) and use a specific nonlinear interpolation instead of the linear interpolation for $\x_t$,
Eq.~\eqref{eq:objective-x} becomes the weighted denoising objective of the diffusion model~\citep{vincent2011connection}, and Eq.~\eqref{eq:ode} becomes the probability flow ODE (PF-ODE)~\citep{song2020score}.

\subsection{Reflow}

\begin{algorithm}[H]
    \caption{Reflow Procedure}
    \label{algo:reflow}
    \begin{algorithmic}[1]
        \State \textbf{First iteration:} 
        \State \quad $\thetab_1 = \arg \min_{\thetab} \mathbb{E}_{\x, \z \sim p^0_{\x \z}} \mathbb{E}_{t \sim p_t} \left[\omega(t) \|\x - \x_{\thetab} (\x_t, t)\|_2^2 \right]$
        \Comment{Train 1-rectified flow}
    \For{$k = 1$ to $K-1$}
        \State $T^{k}(\z) = \z + \int_1^0 \frac{1}{t} \left(\z_t - \x_{\thetab_{k}}(\z_t, t)\right) dt$ with $\z_1 = \z$
        \State $p^{k}_{\x \z}(\x, \z) = p_\z(\z) \delta\left(\x - T^{k}(\z)\right)$ 
        \Comment{Generate synthetic pairs for next coupling}
        \State $\thetab_{k+1} = \arg \min_{\thetab} \mathbb{E}_{\x, \z \sim p^k_{\x \z}} \mathbb{E}_{t \sim p_t} \left[\omega(t) \|\x - \x_{\thetab} (\x_t, t)\|_2^2 \right]$
        \Comment{Train $(k+1)$-rectified flow}
    \EndFor
    \end{algorithmic}
\end{algorithm}

The independent coupling $p_{\x \z}(\x,\z) = p_{\x}(\x) p_{\z} (\z)$ is known to lead to curved ODE trajectories, which require a large number of function evaluations (NFE) to generate high-quality samples~\citep{pooladian2023multisample,lee2023minimizing}.
Reflow~\citep{liu2022flow} is a recursive training algorithm to find a better coupling that yields straighter ODE trajectories. 
Starting from the independent coupling $p^0_{\x \z}(\x,\z) = p_{\x}(\x) p_{\z} (\z)$, the Reflow algorithm generates $p^{k+1}_{\x \z}(\x,\z)$ from $p^{k}_{\x \z}(\x,\z)$ by first generating synthetic $(\x,\z)$ pairs from $p^k_{\x \z}$, then training rectified flow on the generated synthetic pairs (Figure \ref{fig:rectified_flow}$(b)-(d)$). 
We call the vector field resulting from the $k$-th iteration of this procedure $k$-rectified flow.
Pseudocode for Reflow is provided in Algorithm.~\ref{algo:reflow}.

\noindent \textbf{Convergence:} 
\citet{liu2022flow} show that Reflow trajectories are straight in the limit as $K \rightarrow \infty$~.
Hence, to achieve perfectly straight ODE paths that allow for accurate one-step generation, Reflow may need to be applied many times until equilibrium, with each training stage requiring many data-noise pairs, training the model until convergence, and a degradation in generated sample quality.
\textbf{Prior work has empirically found that  Reflow should be applied at least twice} (i.e. $3$-rectified flow) for reasonably good one-step generative performance~\citep{liu2022flow,liu2023instaflow}.
This has been a major downside for rectified flows compared to knowledge distillation methods, which typically require only one distillation stage~\citep{luhman2021knowledge,song2023consistency,zheng2022fast}.

\section{Applying Reflow Once is Sufficient}
\label{sec:claim}

In this section, we argue that under practical settings, the trajectory curvature of the optimal 2-rectified flow is actually close to zero. Hence, prior empirical results requiring more rounds of Reflow may be the result of suboptimal training techniques, and we should focus on improving those training techniques rather than stacking additional Reflow stages.

First, note that the curvature of the optimal 2-rectified flow is zero if and only if the linear interpolation trajectories of 1-rectified flow-generated pairs do not intersect, or equivalently, $\mathbb E[\x | \x_t = (1-t)\x' + t\z'] = \x'$ for all pairs $(\x', \z')$~\citep{liu2022flow}.

\begin{figure}[t]
    \centering
    \includegraphics[width=0.9\linewidth]{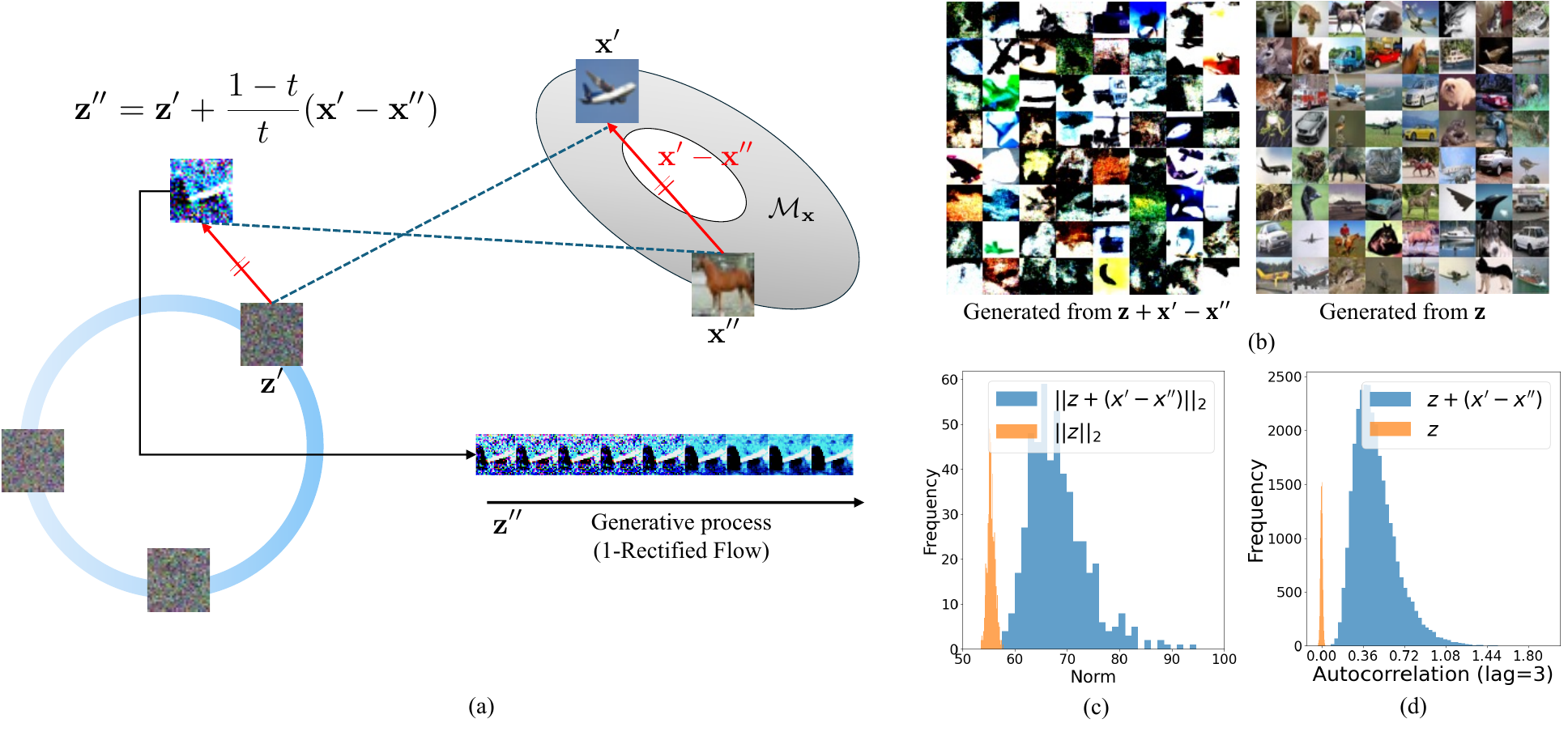}
    \caption{An illustration of the intuition in Sec.~\ref{sec:claim}.
    (a) If two linear interpolation trajectories intersect, $\z'' - \z'$ is parallel to $\x' - \x''$.
    This generally maps $\z''$ to an atypical (e.g., one with high autocorrelation or a norm that is too large to be on a Gaussian annulus)
    realization of Gaussian noise, so the 1-rectified flow cannot reliably map $\z''$ to $\x''$ on $\mathcal M_{\x}$.
    (b) Generated samples from the pre-trained 1-rectified flow starting from $\z\sim \mathcal N(\zerob,\Ib)$ (\textit{right}), which is the standard setting, and $\z ''=\z + (\x' - \x'')$, where $\x',\x''$ are sampled from 1-rectified flow trained on CIFAR-10 (\textit{left}). Qualitatively, we see that the left samples have very low quality.
    (c) Empirically, we show the $\ell_2$ norm of $z''=\z + (\x' - \x'')$ compared to $z'$, which is sampled from the standard Gaussian. $\z''$ generally lands outside the annulus of typical Gaussian noise.
    (d) $\z + (\x' - \x'')$ has high autocorrelation while the autocorrelation of Gaussian noise is nearly zero in high-dimensional space. 
    }
    \label{fig:claim}
  \end{figure}

To begin with, consider the manifold $\mathcal M_{\x}$ of the synthetic distribution $p^1(\x) = \int p^1(\x, \z) d\z$. Consider two points $\x'$ and $\x''$ from the manifold, and two noises $\z'$ and $\z''$ that are mapped to $\x'$ and $\x''$ by 1-rectified flow.
Here, we say two pairs $(\x', \z')$ and $(\x'', \z'')$ intersect if  $\exists t\in [0,1]$ s.t. $(1-t)\x' + t\z' = (1-t)\x'' + t\z''$. For example, in Figure \ref{fig:claim}(a), we observe that the two trajectories intersect at an intermediate $t$.

For an intersection to exist at $t$ it must hold that 1) 1-rectified flow  maps $\z''$ to $\x''$, and 2)  $\z''=\z' + \frac{1-t}{t}(\x' - \x'')$ by basic geometry.
However, note that for realistic data distributions and if 1-rectified flow is sufficiently well-trained, $\z''=\z' + \frac{1-t}{t}(\x' - \x'')$ is not a common noise realization (e.g., it is likely to have nonzero autocorrelation or a norm that is too large to be on a Gaussian annulus), as shown visually in Figure \ref{fig:claim}(a).
As 1-rectified flow is almost entirely trained on common Gaussian noise inputs, 
it cannot generally map an atypical $\z''$ to $\mathcal M_{\x}$. 
Figure \ref{fig:claim}(c) shows qualitatively that if we draw values of  $\z''$ by first drawing $\z'\sim \mathcal N(0,I)$ and then adding $(\x'-\x'')$ for independent draws of $\x',\x''$, the $\z''$ vectors fall outside the annulus of typical standard Gaussian noise. 
Similarly, Figure \ref{fig:claim}(d) shows that the constructed noise vectors $\z''$ have higher autocorrelation than expected.
As a result, Figure \ref{fig:claim}(b) visually shows that the generated samples have little overlap with the expected samples from typical draws of $\z'$.

This suggests empirically that when training 2-rectified flow, intersections are rare (i.e. $\mathbb E[\x | \x_t = (1-t)\x' + t\z'] \approx \x'$), which in turn implies that the optimal 2-rectified flow trajectories are nearly straight.
Hence, additional rounds of Reflow are unnecessary, while also degrading sample quality.
This intuition allows us to focus on better training techniques for 2-rectified flow rather than training 3- or 4-rectified flow. It also leads us to several improved techniques, discussed in Sec.~\ref{sec:techniques}.

\textbf{Edge cases:} Note that if $||\x' - \x''||_2$ is small, 1-rectified flow could map $\z''$ to some point on $\mathcal M_{\x}$. However, it does not alter the conclusion because the average of $\x'$ and $\x''$ is close to $\x'$ anyway, so 
$\mathbb E[\x | \x_t = (1-t)\x' + t\z'] \approx \x'$.
  Similarly, if $t$ is close to 1, $\frac{1-t}{t}(\x' - \x'') \approx \zerob$, so 1-rectified flow can map $\z''$ to $\mathcal M_{\x}$. If the 1-rectified flow is $L$-Lipschitz, 
  $||\x' - \x''||_2 \leq L ||\z' - \z''||_2$.
  Therefore, the expectation $\mathbb E[\x | \x_t ]$ again will not deviate much from $\x'$.
\section{Improved Training Techniques for Reflow}
\label{sec:techniques}
The observation in Sec.~\ref{sec:claim} suggests that the optimal 2-rectified flow is nearly straight.
Therefore, if the one-step generative performance of the 2-rectified flow model is not as good as expected, it is likely due to suboptimal training.
In this section, we show that the few-step generative performance of the 2-rectified flow can be significantly improved by applying several new training techniques.

\subsection{Timestep distribution}
\label{sec:timestep}

As in diffusion models, rectified flows are trained on randomly sampled timesteps $t$, and the distribution from which $t$ is sampled is an important design choice.
Ideally, we want to focus the training effort on timesteps that are more challenging rather than wasting computational resources on easy tasks.
One common approach is to focus on the tasks where the training loss is high~\citep{shrivastava2016training}.
However, the training error of rectified flows is not a reliable measure of difficulty because different timesteps have different non-zero lower bounds.
To understand this, let us decompose the training error into two terms:
\begin{align}
    \mathcal L(\thetab,t) := \frac{1}{t^2}\mathbb E [||\x - \x_{\thetab} (\x_t, t)||_2^2] = \underbrace{\frac{1}{t^2} \mathbb E [||\x - \mathbb E[\x | \x_t]||_2^2]}_{\text{Lower bound}} + \mathcal {\bar L}_{\text{}}(\thetab, t).
    \label{eq:loss-decomposition}
\end{align}

\begin{wrapfigure}[]{r}{0.45\textwidth}
    \centering
    \includegraphics[width=0.4\textwidth]{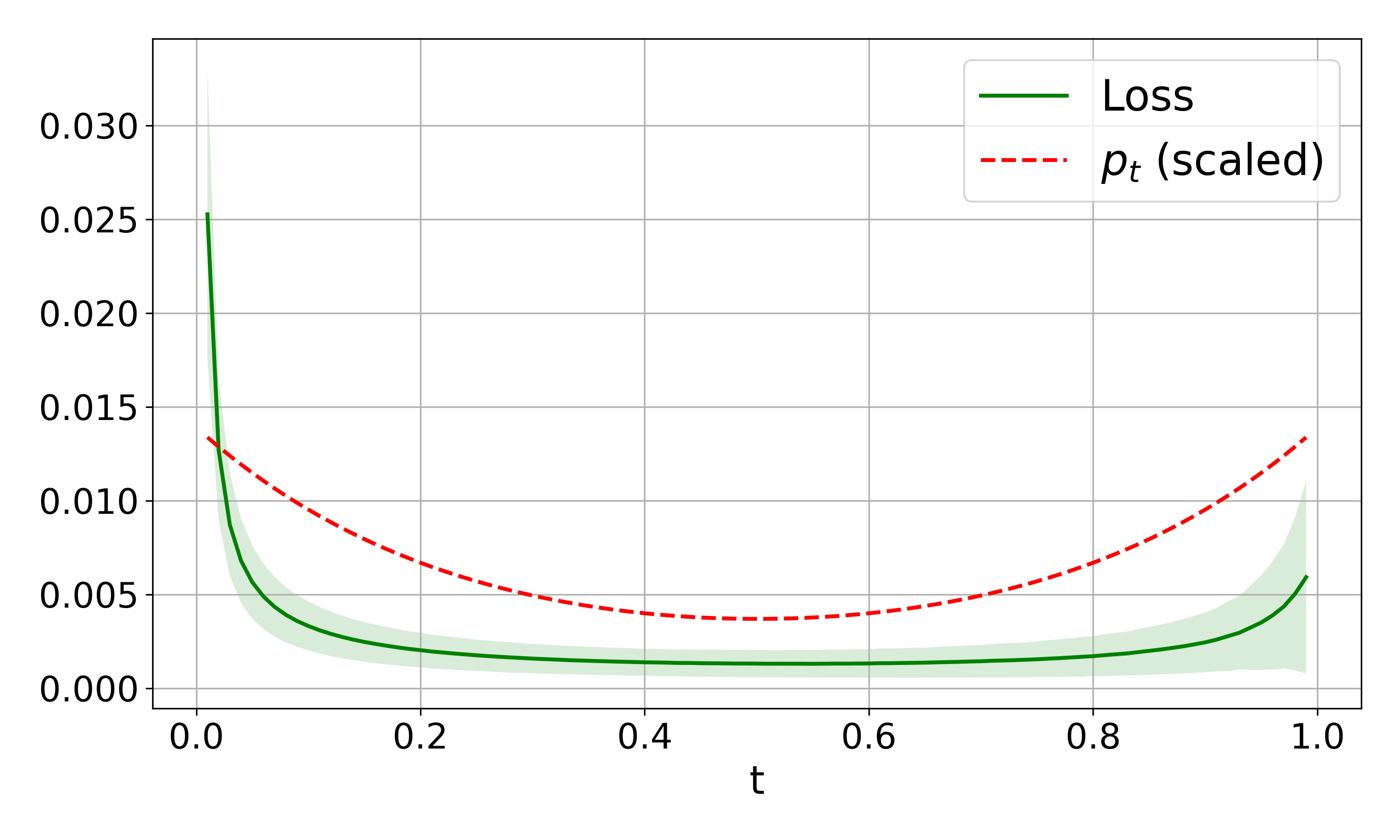}
    \vspace{-5mm}
    \caption{Training loss of the vanilla 2-rectified flow on CIFAR-10 measured on $5,000$ samples after $200,000$ iterations.
    The shaded area represents the 1 standard deviation of the loss.
    The dashed curve is our U-shaped timestep distribution, scaled by a constant factor for visualization. }
    \vspace{-6mm}
    \label{fig:loss}
\end{wrapfigure}

The first term does not depend on $\thetab$ and thus cannot be reduced.
The second term represents the actual minimizable training error, but its value cannot be directly observed because the first term is usually unknown.
Fortunately, because of the finding in Sec.~\ref{sec:claim}, we expect that the first term is nearly zero when training 2-rectified flow, so we can use $\mathcal L(\thetab,t)$ for designing the timestep distribution.

Figure~\ref{fig:loss} shows that the training loss 
of 2-rectified flow is large at each end of the interval $t \in [0, 1]$ and small in the middle.
We thus propose to use a U-shaped timestep distribution for $p_t$. 
Specifically, we define $p_t(u) \propto \exp(au) + \exp(-au)$ on $u \in [0, 1]$.
We find that $a=4$ works well in practice (Table~\ref{tab:ablation}).
Compared to the uniform timestep distribution (config B), the U-shaped distribution (config C) improves the FID of 2-rectified flow from $7.14$ to $5.17$ (a $28\%$ improvement) on CIFAR-10, $12.39$ to $9.03$ ($27\%$) on AFHQ, and $8.84$ to $6.81$ ($23\%$) on FFHQ in the one-step setting.

\begin{table}[]
\footnotesize
    \caption{Effects of the improved training techniques. 
    The baseline (config A) is the 2-rectified flow with the uniform timestep distribution and the squared $\ell_2$ metric~\citep{liu2022flow}.
    Config B is the improved baseline with EDM initialization (Sec.~\ref{sec:init}) and increased batch size ($128 \rightarrow 512$ on CIFAR-10).
    FID (the lower the better) is computed using $50,000$ synthetic samples and the entire training set.
    We train the models for $800,000$ iterations on CIFAR-10 and $1,000,000$ iterations on AFHQ and FFHQ and report the best FID for each setting.
    }

    \centering
    \begin{tabular}{@{}lllllll@{}}
    \toprule
    & \multicolumn{2}{c}{CIFAR-10} & \multicolumn{2}{c}{AFHQ $64 \times 64$} & \multicolumn{2}{c}{FFHQ $64 \times 64$} \\  \midrule
    Base~\citep{liu2022flow} (\textsc{a})              & 12.21          & -          & -     & -       & -       & -       \\
    (\textsc{a}) + EDM init + larger batch size (\textsc{b})              & 7.14          & 3.61          & 12.39      & 4.16       & 8.84       & 4.79       \\
    (\textsc{b}) + Our $p_t$ (\textsc{c})        & 5.17          & 3.37          & 9.03       & 3.61       & 6.81       & 4.66       \\
    (\textsc{c}) + Huber (\textsc{d})          & 5.24          & 3.34          & 8.20       & 3.55       & 7.06       & 4.79       \\
    (\textsc{c}) + LPIPS-Huber (\textsc{e})     & 3.42          & 2.95          & 4.13       & 3.15       & \textbf{5.21}       & \textbf{4.26}       \\
    (\textsc{c}) + LPIPS-Huber-$\frac{1}{t}$ (\textsc{f}) & {3.38}       & {2.76}       & \textbf{4.11}       & \textbf{3.12}       & 5.65       & 4.41       \\ 
    (\textsc{f}) + Incorporating real data (\textsc{g}) & \textbf{3.07}       & \textbf{2.40}       & -       & -       & -       & -       \\
    \midrule
    NFE & \multicolumn{1}{c}{1} & \multicolumn{1}{c}{2} & \multicolumn{1}{c}{1} & \multicolumn{1}{c}{2} & \multicolumn{1}{c}{1} & \multicolumn{1}{c}{2} \\ \bottomrule
    \end{tabular}
    \label{tab:ablation}
\end{table}

For 1-rectified flow training, $p_t$ was chosen to be the uniform distribution~\citep{liu2022flow,liu2023instaflow} or logit-normal distribution~\citep{esser2024scaling} which puts more emphasis on the middle of the interval.
When training 1-rectified flow, a model learns to simply output the dataset average when $t=1$ and the noise average (i.e., zero) when $t=0$.
The meaningful part of the training thus happens in the middle of the interval.
In contrast, from Eq.~\eqref{eq:objective-v} we can see that 2-rectified flow learns to directly predict the data from the noise at $t=1$ and the noise from the data at $t=0$, which are nontrivial tasks.
Therefore, the U-shaped timestep distribution is more suitable for 2-rectified flow.

\subsection{Loss function}
\label{sec:metrics}
Previously, the squared $\ell_2$ distance was used as the training metric for rectified flow to obtain the MMSE estimator $\mathbb E[\x | \x_t]$.
However, as we have shown in Sec.~\ref{sec:claim} that $\mathbb E[\x | \x_t = (1-t)\x' + t\z'] \approx \x'$, we can generalize Eq.~\eqref{eq:objective-x} or equivalently Eq.~\eqref{eq:objective-v} to any premetric $m$ (i.e. $m(\ab, \bb) = 0 \Leftrightarrow \ab = \bb$):
\begin{align}
    \min_{\thetab} \mathbb E_{\x, \z \sim p_{\x \z}} \mathbb E_{t \sim p_t} [m(\z - \x, \vb_{\thetab}(\x_t, t))],
    \label{eq:objective-m}
\end{align}
Note that \textbf{without the intuition in Sec.~\ref{sec:claim}, only the squared $\ell_2$ distance would have been a valid premetric}, as any other premetric makes the model deviate from the intended optimum (the posterior expectation $\mathbb E[\x | \x_t]$).
Although the choice of $m$ does not affect the optimum, it does affect the training dynamics and thus the obtained model. Other than the squared $\ell_2$ distance, we consider the following premetrics:
\begin{itemize}[leftmargin=*]
    \item Pseudo-Huber~\citep{charbonnier1997deterministic,song2023improved}: $m_{\text{hub}}(\z - \x, \vb_{\thetab}((\x_t, t)) = \sqrt{||\z - \x - \vb_{\thetab}(\x_t, t)||_2^2 + c^2} - c$, where $c = 0.00054{d}$ with $d$ being data dimensionality.
    \item LPIPS-Huber: $m_{\text{lp-hub}}(\z - \x, \vb_{\thetab}(\x_t, t)) = (1-t) m_{\text{hub}}(\z - \x, \vb_{\thetab}(\x_t, t)) + \text{LPIPS}(\x, \x_t - t \cdot \vb_{\thetab}(\x_t, t))$, where LPIPS$(\cdot, \cdot)$ is the learned perceptual image patch similarity~\citep{zhang2018unreasonable}.
    \item LPIPS-Huber-$\frac{1}{t}$: $m_{\text{lp-hub-}\frac{1}{t}}(\z - \x, \vb_{\thetab}(\x_t, t)) = (1-t) m_{\text{hub}}(\z - \x, \vb_{\thetab}(\x_t, t)) + \frac{1}{t}\text{LPIPS}(\x, \x_t - t \cdot \vb_{\thetab}(\x_t, t))$,
\end{itemize}

The Pseudo-Huber loss is less sensitive to the outliers than the squared $\ell_2$ loss, which can potentially reduce the gradient variance~\citep{song2023improved} and make training easier.
In our initial experiments, we found that the Pseudo-Huber loss tends to work better than the squared $\ell_2$ loss with a small batch size (e.g. 128 on CIFAR-10).
When the batch size is sufficiently large, it performs on par with the squared $\ell_2$ loss on CIFAR-10 and FFHQ-64 and outperforms it on AFHQ-64, as shown in Table~\ref{tab:ablation}.
As it is less sensitive to the batch size, we choose to use the Pseudo-Huber loss in the following experiments.

We also explore the LPIPS, which forces the model to focus on reducing the perceptual distance between the generated data and the ground truth. 
Since LPIPS is not a premetric as two different points could have zero LPIPS if they are perceptually similar, we use it in combination with the Pseudo-Huber loss with the weighting $1-t$,
thereby relying more on LPIPS when $t$ is close to 1 where the task is more challenging.
Note that in $m_{\text{lp-hub}}$, the gradient vanishes when $t$ is close to zero.
To compsensate, we experiment with $m_{\text{lp-hub-}\frac{1}{t}}$ where we multiply LPIPS by $\frac{1}{t}$.
Compared to config D, the LPIPS-Huber loss improves the FID of 2-rectified flow from $5.24$ to $3.38$ (a $35\%$ improvement) on CIFAR-10, $8.20$ to $4.11$ ($50\%$) on AFHQ, and $7.06$ to $5.21$ ($26\%$)  on FFHQ in the one-step setting, as seen in Table~\ref{tab:ablation}.

\subsection{Initialization with pre-trained diffusion models}
\label{sec:init}
Training 1-rectified flow from scratch is computationally expensive. Recently, \citet{pokle2023training} showed that pre-trained diffusion models can be used to approximate $\mathbb E[\x | \x_t = \z_t]$ in Eq.~\eqref{eq:ode} by adjusting the signal-to-noise ratio.
The following proposition is the special cases of Lemma 2 of \citet{pokle2023training} restated with extended proof and a minor fix. We provide the constants and proof in Appendix.~\ref{sec:proof-prop}.

\begin{proposition}
    Let $p^{\text{RE}}(\x | \x_t, t)$ be the posterior distribution of the perturbation kernel $\mathcal N((1-t) \x, t^2\Ib)$. 
    Also, let $p^{\text{VP}}(\x | \x_t, t)$ and $p^{\text{VE}}(\x | \x_t, t)$ be the posterior distributions of $\mathcal N(\alpha(t) \x, (1-\alpha(t))^2 \Ib)$ and $\mathcal N(\x, t^2 \Ib)$, each.
    Then, 
    \begin{align}
        \int p^{\text{RE}}(\x | \x_t = \z_t, t) \x \  d\x = \int p^{\text{VP}}(\x | \x_t = s_{\text{VP}} \z_t, t_{\text{VP}}) \x \ d\x = \int p^{\text{VE}}(\x | \x_t = s_{\text{VP}} \z_t, t_{\text{VE}}) \x \ d\x,
        \label{eq:three-integrals}
    \end{align}
    where $s_{\text{VP}}$ and $s_{\text{VE}}$ are the scaling factors and $t_{\text{VP}}$ and $t_{\text{VE}}$ are the converted times for the VP and VE diffusion models, respectively.
    \label{prop:conversion}
\end{proposition}

\begin{table}[t]
    \caption{The converted time and scale for the variance preserving (VP) and variance exploding (VE) diffusion models.
   Here, $\alpha(t) = \exp(-\frac{1}{2} \int_0^t (19.9s + 0.1) ds)$ following \citet{song2020score}, and the perturbation kernel of the VE diffusion is $\mathcal N(\x, t^2 \Ib)$ following \citet{karras2022elucidating}.
   }
    \center
  \begin{tabular}{cccc}
  \toprule
  $t_{\text{VP}}$ & $t_{\text{VE}}$ & $s_{\text{VP}}$ & $s_{\text{VE}}$ \\
  \midrule
  $\frac{1}{9.95}\left (-0.05 + \sqrt{0.0025 - 19.9 \cdot \ln   \frac{1-t}{\sqrt{(1-t)^2 + t^2}}  }\right )$ & $\frac{t}{1-t}$ & $\frac{\alpha(t_{\text{VP}})}{1-t}$ & $\frac{1}{1-t}$ \\
  \bottomrule
  \end{tabular}
   \label{tab:time-scale}
\end{table}

We have explicitly computed the time and scale conversion factors for the VP and VE diffusion models in Table~\ref{tab:time-scale}. See Appendix \ref{app:initialization_results} for derivation.

Proposition~\ref{prop:conversion} allows us to initialize the Reflow with the pre-trained diffusion models such as EDM~\citep{karras2022elucidating} or DDPM~\citep{ho2020denoising} and use Table~\ref{tab:time-scale} to adjust the time and scaling factors.

Starting from the vanilla 2-rectified flow setup~\citep{liu2022flow} (config A), we initialize 1-rectified flow with the pre-trained EDM (VE).
We also increase the batch size from $128$ to $512$ on CIFAR-10 compared to \citet{liu2022flow}. 
Overall, these improve the FID of 2-rectified flow from $12.21$ to $7.14$ (a $42\%$ improvement) in the one-step setting on CIFAR-10 (config B).

\subsection{Incorporating real data}
Training 2-rectified flow does not require real data (i.e., it can be data-free), but we can use real data if it is available.
To see the effects of incorporating real data, we integrate the generative ODE of 1-rectified flow backward from $t=0$ to $t=1$ using an NFE of 128 to collect $50,000$ pairs of (real data, synthetic noise) on CIFAR-10. For quick validation, we take the pre-trained 2-rectified flow model (config F) and fine-tune it using the (real data, synthetic noise) pairs for $5,000$ iterations with a learning rate of 1e-5. This improves the FID of 2-rectified flow from $3.38$ to $3.07$ in the one-step setting on CIFAR-10 (config G).

In this fine-tuning setting, we also explored using (synthetic data, real noise) pair with a probability of $p$, but we found that not incorporating (synthetic data, real noise) pairs at all (i.e., $p=0$) performs the best. We expect that training from scratch will further improve the performance with different values of $p$, and leave it to future work. A similar idea is also explored in \citet{anonymous2024balanced}.

\begin{table*}
\small
    \begin{minipage}[t]{0.49\linewidth}
	\caption{Unconditional generation on CIFAR-10.}
    \label{tab:cifar-10}
	\centering
	{\setlength{\extrarowheight}{0.4pt}
	\begin{adjustbox}{max width=\linewidth}
	\begin{tabular}{@{}l@{\hspace{-0.2em}}c@{\hspace{0.3em}}c@{\hspace{0.3em}}c@{}}
        \Xhline{3\arrayrulewidth}
	    METHOD & NFE ($\downarrow$) & FID ($\downarrow$) & IS ($\uparrow$) \\
        \\[-2ex]
        \multicolumn{4}{@{}l}{\textbf{Diffusion models}}\\\Xhline{3\arrayrulewidth}
        Score SDE \citep{song2020score} & 2000 & 2.38 & 9.83\\
        DDPM \citep{ho2020denoising} & 1000 & 3.17 & 9.46\\
        LSGM \citep{vahdat2021score} & 147 & 2.10 &\\
        EDM \citep{karras2022elucidating}
         & 35 & 1.97 &  \\
        \multicolumn{4}{@{}l}{\textbf{Distilled diffusion models}}\\\Xhline{3\arrayrulewidth}
        Knowledge Distillation \citep{luhman2021knowledge} & 1 & 9.36 &  \\
        DFNO (LPIPS) \citep{zheng2022fast} & 1 & 3.78 & \\
        TRACT \citep{berthelot2023tract} & 1 & 3.78 & \\
         & 2 & \textcolor{blue}{3.32} & \\
        PD \citep{salimans2022progressive} & 1 & 9.12 &  \\
          & 2 & 4.51 &  \\
        \multicolumn{4}{@{}l}{\textbf{Score distillation}}\\\Xhline{3\arrayrulewidth}
        Diff-Instruct \citep{luo2024diff} & 1 & 4.53 & 9.89\\
        DMD \citep{yin2023one} & 1 & 3.77 & \\
        \multicolumn{4}{@{}l}{\textbf{GANs}}\\\Xhline{3\arrayrulewidth}
        BigGAN \citep{brock2018large} & 1 & 14.7 & 9.22\\
        StyleGAN2 \citep{karras2020analyzing}
         & 1 & 8.32 & 9.21\\
        StyleGAN2-ADA \citep{karras2020training}
         & 1 & \textcolor{red}{2.92} & 9.83\\
        \multicolumn{4}{@{}l}{\textbf{Consistency models}}\\\Xhline{3\arrayrulewidth}
        CD (LPIPS) \citep{song2023consistency} & 1 & 3.55 & 9.48 \\
          & 2 & \textcolor{blue}{2.93} & 9.75 \\
        CT (LPIPS) \citep{song2023consistency} & 1 & 8.70 & 8.49 \\
          & 2 & 5.83 & 8.85 \\
        iCT \citep{song2023improved} & 1 & \textcolor{red}{2.83} & 9.54 \\
        & 2 & \textcolor{blue}{2.46} & 9.80 \\
        iCT-deep \citep{song2023improved} & 1 & \textcolor{red}{2.51} & 9.76 \\
        & 2 & \textcolor{blue}{\textbf{2.24}} & 9.89 \\
        CTM \citep{kim2023consistency} & 1 & 5.19 &  \\
        CTM \citep{kim2023consistency} + GAN & 1 & \textcolor{red}{\textbf{1.98}} &  \\
        \multicolumn{4}{@{}l}{\textbf{Rectified flows}}\\\Xhline{3\arrayrulewidth}
        1-rectified flow (+distill) \citep{liu2022flow}
         & 1 & 6.18 & 9.08\\
        2-rectified flow  \citep{liu2022flow}
         & 1 & 12.21 & 8.08\\
         & 110 & 3.36 & 9.24 \\
        +distill \citep{liu2022flow}
         & 1 & 4.85 & 9.01\\
        3-rectified flow  \citep{liu2022flow}
         & 1 & 8.15 & 8.47 \\
         & 104 & 3.96 & 9.01 \\
        +Distill \citep{liu2022flow}
         & 1 & 5.21 & 8.79 \\
          
        \textbf{2-rectified flow++ (ours)} & 1 & \textcolor{red}{3.07} &  \\
        & 2 & \textcolor{blue}{2.40} &  \\
	\end{tabular}
    \end{adjustbox}
	}
\end{minipage}
\hfill
\begin{minipage}[t]{0.49\linewidth}
    \caption{Class-conditional generation on ImageNet $64\times 64$.}
    \label{tab:imagenet-64}
    \centering
    {\setlength{\extrarowheight}{0.4pt}
    \begin{adjustbox}{max width=\linewidth}
    \begin{tabular}{@{}l@{\hspace{0.2em}}c@{\hspace{0.3em}}c@{\hspace{0.3em}}c@{\hspace{0.3em}}c@{}}
        \Xhline{3\arrayrulewidth}
        METHOD & NFE ($\downarrow$) & FID ($\downarrow$) & Prec. ($\uparrow$) & Rec. ($\uparrow$) \\
        \\[-2ex]
        \multicolumn{4}{@{}l}{\textbf{Diffusion models}}\\\Xhline{3\arrayrulewidth}
        DDIM \citep{song2020denoising} & 50 & 13.7 & 0.65 & 0.56\\
        & 10 & 18.3 & 0.60 & 0.49\\
        DPM solver \citep{lu2022dpm} & 10 & 7.93 &&\\
        & 20 & 3.42 &&\\
        DEIS \citep{zhang2022fast} & 10 & 6.65 & \\
        & 20 & 3.10 && \\
        DDPM \citep{ho2020denoising} & 250 & 11.0 & 0.67 & 0.58 \\
        iDDPM \citep{nichol2021improved} & 250 & 2.92 & 0.74 & 0.62\\
        ADM \citep{dhariwal2021diffusion} & 250 & 2.07 & 0.74 & 0.63\\
        EDM \citep{karras2022elucidating} & 79 & 2.30 &  & \\

        \multicolumn{4}{@{}l}{\textbf{Distilled diffusion models}}\\\Xhline{3\arrayrulewidth}
        DFNO (LPIPS) \citep{zheng2022fast} & 1 & 7.83 & & 0.61\\
        TRACT \citep{berthelot2023tract} & 1 & 7.43 & & \\
            & 2 & 4.97 & & \\
        BOOT \citep{gu2023boot} & 1 & 16.3 & 0.68 & 0.36\\
        PD \citep{salimans2022progressive} & 1 & 15.39 & 0.59 & 0.62 \\
            & 2 & 8.95 & 0.63 & 0.65 \\
            & 4 & 6.77 & 0.66 & 0.65 \\

        \multicolumn{4}{@{}l}{\textbf{Score distillation}}\\\Xhline{3\arrayrulewidth}
        Diff-Instruct \citep{luo2024diff} & 1 & 5.57 & &\\
        DMD \citep{yin2023one} & 1 & \textcolor{red}{2.62} & &\\

        \multicolumn{4}{@{}l}{\textbf{GANs}}\\\Xhline{3\arrayrulewidth}
        BigGAN-deep \citep{brock2018large} & 1 & \textcolor{red}{4.06} & 0.79 & 0.48\\
        \multicolumn{4}{@{}l}{\textbf{Consistency models}}\\\Xhline{3\arrayrulewidth}
        CD (LPIPS) \citep{song2023consistency} & 1 & 6.20 & 0.68 & 0.63 \\
            & 2 & \textcolor{blue}{4.70} & 0.69 & 0.64 \\
            & 3 & 4.32 & 0.70 & 0.64 \\
        CT (LPIPS) \citep{song2023consistency} & 1 & 13.0 & 0.71 & 0.47 \\
            & 2 & 11.1 & 0.69 & 0.56\\
        iCT \citep{song2023improved}  & 1 & \textcolor{red}{4.02} & 0.70 & 0.63\\
        & 2 & \textcolor{blue}{3.20} & 0.73 & 0.63 \\
        iCT-deep \citep{song2023improved}  & 1 & \textcolor{red}{3.25} & 0.72 & 0.63 \\
        & 2 & \textcolor{blue}{2.77} & 0.74 & 0.62 \\
        CTM + GAN \citep{kim2023consistency} & 1 & \textcolor{red}{\textbf{1.92}} &  & 0.57 \\
         & 2 & \textcolor{blue}{\textbf{1.73}} &  & 0.57 \\
        \multicolumn{4}{@{}l}{\textbf{Rectified flows}}\\\Xhline{3\arrayrulewidth}
        \textbf{2-rectified flow++ (ours)} & 1 & 4.31 &  &  \\
         & 2 & \textcolor{blue}{3.64} &  &  \\
    \end{tabular}
    \end{adjustbox}
    }
\end{minipage}
\captionsetup{labelformat=empty, labelsep=none, font=scriptsize}
\caption{The \textcolor{red}{red} rows correspond to the top-5 baselines for the 1-NFE setting, and the \textcolor{blue}{blue} rows correspond to the top 5 baselines for the 2-NFE setting. The lowest FID scores for 1-NFE and 2-NFE are \textbf{boldfaced}.}
\end{table*}

\section{Experiments}
\label{sec:exp}
We call these combined improvements to Reflow training \emph{2-rectified flow++} and evaluate it on four datasets: 
CIFAR-10 \cite{krizhevsky2009learning}, AFHQ \cite{choi2020stargan}, FFHQ \cite{karras2019style}, and ImageNet \cite{deng2009imagenet}.
We compare our improved Reflow to up to 20 recent baselines, in the families of diffusion models, distilled diffusion models, score distillation, GANs, consistency models, and rectified flows.
The details of our experimental setup are included in Appendix~\ref{sec:exp-details}.

\subsection{Unconditional and class-conditional image generation}
\label{sec:unconditional}

In Tables~\ref{tab:cifar-10} and~\ref{tab:imagenet-64}, we compare 2-rectified flow++ with the state-of-the-art methods on CIFAR-10 and ImageNet $64 \times 64$.
We observe two main messages:

\textbf{On both datasets, 2-rectified flow++ (ours) outperforms or is competitive with SOTA baselines in the 1-2 NFE regime.}
On CIFAR-10 (Table \ref{tab:cifar-10}), our 2-rectified flow achieves an FID of $3.07$ in one step, surpassing existing distillation methods such as 
consistency distillation (CD)~\citep{song2023consistency}, progressive distillation (PD)~\citep{salimans2022progressive}, diffusion model sampling with neural operator (DSNO)~\citep{zheng2022fast}, and TRAnsitive Closure Time-distillation (TRACT)~\citep{berthelot2023tract}.
On ImageNet $64 \times 64$ (Table \ref{tab:imagenet-64}), our model surpasses the distillation methods such as CD, PD, DFNO, TRACT, and BOOT in one-step generation.
We also close the gap with iCT (4.01 vs 4.31), the state-of-the-art consistency model, even with half the batch size.
Note that on ImageNet, our model does not use real data during training, while consistency training (CT) requires real data. 
We believe we could further reduce the gap by using config G in Tab.~\ref{tab:ablation}.
Uncurated samples of our model are provided in Appendix.~\ref{sec:sample}.

\textbf{2-rectified flow++ reduces the FID of 2-rectified flows by up to 75\%.}
Compared to vanilla rectified flows~\citep{liu2022flow}, our one-step FID on CIFAR-10 is lower than that of the previous 2-rectified flow by $9.14$ (a reduction of 75\%), and of the 3-rectified flow by $5.08$ (see also Table \ref{tab:ablation} for ablations on other datasets).
In addition, it outperforms the previous 2-rectified flow with 110 NFEs using only one step and also surpasses 2-rectified flow $+$ distillation, which requires an additional distillation stage.

\begin{wraptable}{r}{0.5\textwidth}
    \caption{Comparison of the number of forward passes. Reflow uses $395$M forward passes for generating pairs and $1,433.6$M for training.}
    \centering
    \small
    \begin{tabular}{@{}lccc@{}}
        \toprule
        Method & Per iteration & Total     & Rel. total cost  \\ \midrule
        Reflow & $1$           & $1828.6$M & $\times$1 \\
        CD     & $4$           & $5734.4$M & $\times$3.1 \\
        CT     & $2$           & $2867.2$M & $\times$1.5 \\ \bottomrule
        \end{tabular}
    \label{tab:forward-pass}
\end{wraptable}

\subsection{Reflow can be computationally more efficient than other distillation methods}

At first glance, Reflow seems computationally expensive compared to CD and CT as it requires generating synthetic pairs before training.
However, CD requires 4 (1 for student, 1 for teacher, and 2 for Heun's solver) forward passes for each training iteration, and CT requires 2 (1 for student and 1 for teacher) forward passes, while Reflow requires only 1.
For example, in our ImageNet experiment setting, the total number of forward passes for Reflow is $395$M + $1433.6$M = $1828.6$M (395M for generating pairs and 1,433.6M for training),
while the total numbers of forward passes for CD and CT would be $1,433.6 \cdot 4 = 5,734.4$M and $1,433.6 \cdot 2 = 2,867.2$M under the same setting. 
See Table~\ref{tab:forward-pass} for the comparison.
Moreover, generating pairs is a one-time cost since we can reuse the pairs for multiple training runs.

In terms of the storage cost, the synthetic images for ImageNet $64 \times 64$ require 42 GB. For noise, we only store the states of the random number generator, which is negligible.

While these results should be further validated for larger datasets, our results suggest that the fact that Reflow requires generating synthetic pairs does not necessarily make it less computationally efficient than other distillation methods.

\subsection{Effects of samplers}
\label{sec:ode}
\begin{figure}[]
    \centering
    \begin{subfigure}{0.32\linewidth}
    \includegraphics[width=1\linewidth]{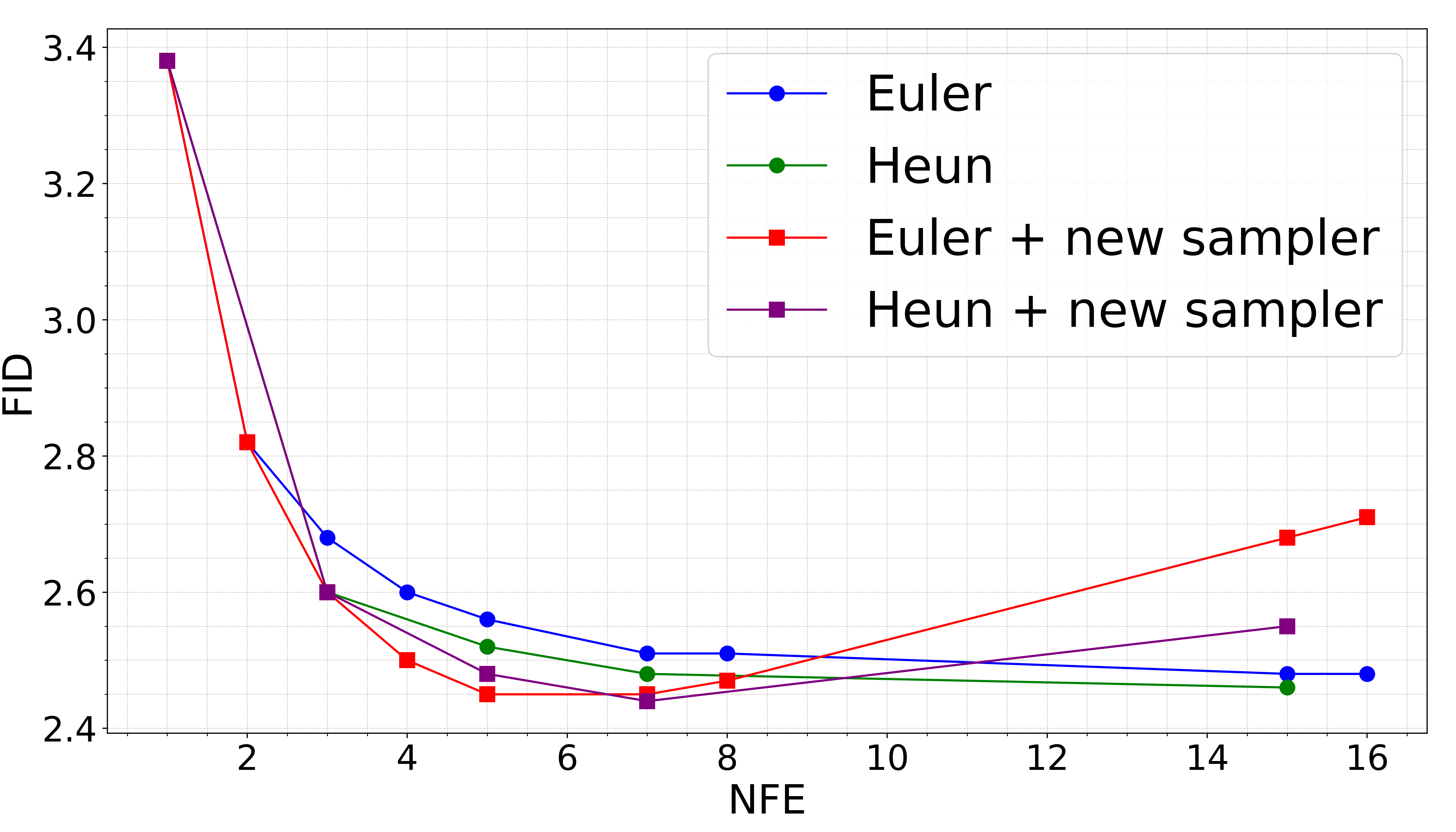}
    \caption{CIFAR-10}
    \end{subfigure}
    \begin{subfigure}{0.32\linewidth}
    \includegraphics[width=1\linewidth]{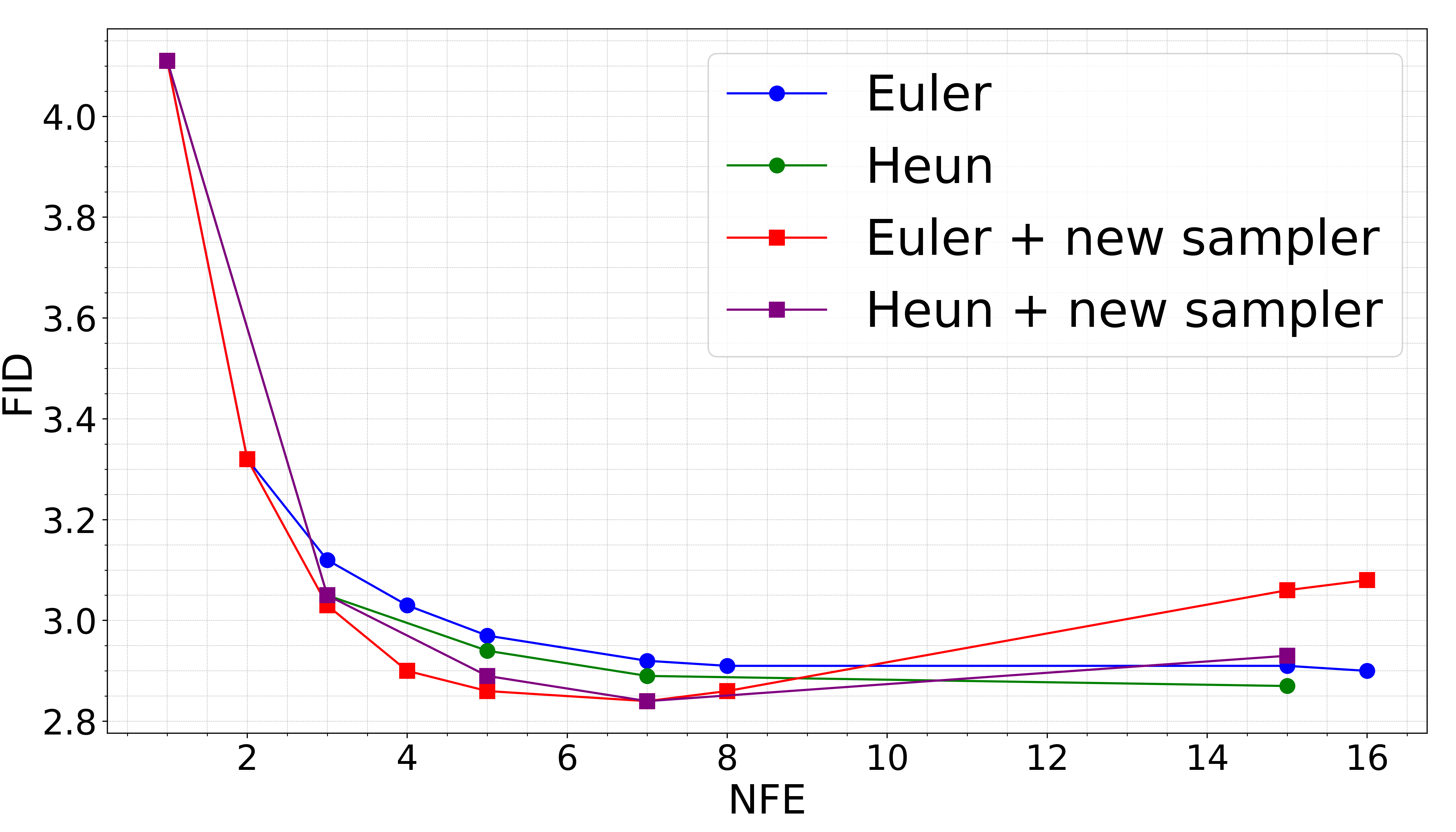}
    \caption{AFHQ $64 \times 64$}
    \end{subfigure}
    \begin{subfigure}{0.32\linewidth}
    \includegraphics[width=1\linewidth]{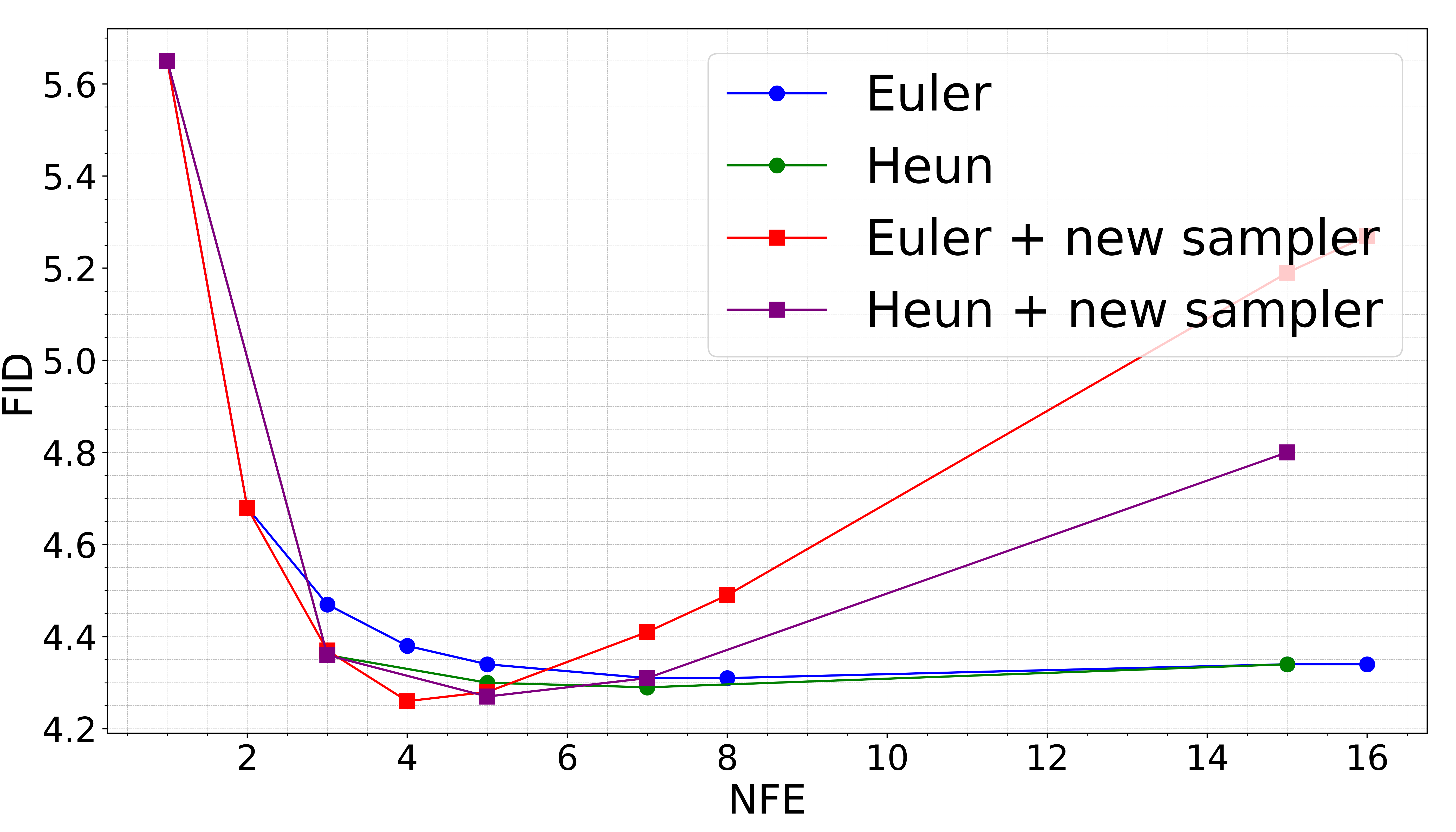}
    \caption{FFHQ $64 \times 64$}
    \end{subfigure}
    \caption{Effects of ODE Solver and new update rule.}
    \label{fig:multi}
\end{figure}

Unlike distillation methods, rectified flow is a neural ODE, and its outputs approach the true solution of the ODE as NFE increases (i.e., precision grows).
Figure~\ref{fig:multi} shows that with the standard Euler solver, FID decreases as NFE increases on all datasets.
Moreover, Heun's second-order solver further improves the trade-off curve between FID and NFE.
This suggests that there may be further room for improvement by using more advanced samplers.
We provide some preliminary ideas towards this goal in Appendix~\ref{sec:pseudo-code}.

\begin{figure}[]
    \centering
    \begin{subfigure}[t]{0.32\linewidth}
        \centering
        \includegraphics[height=0.65\textwidth]{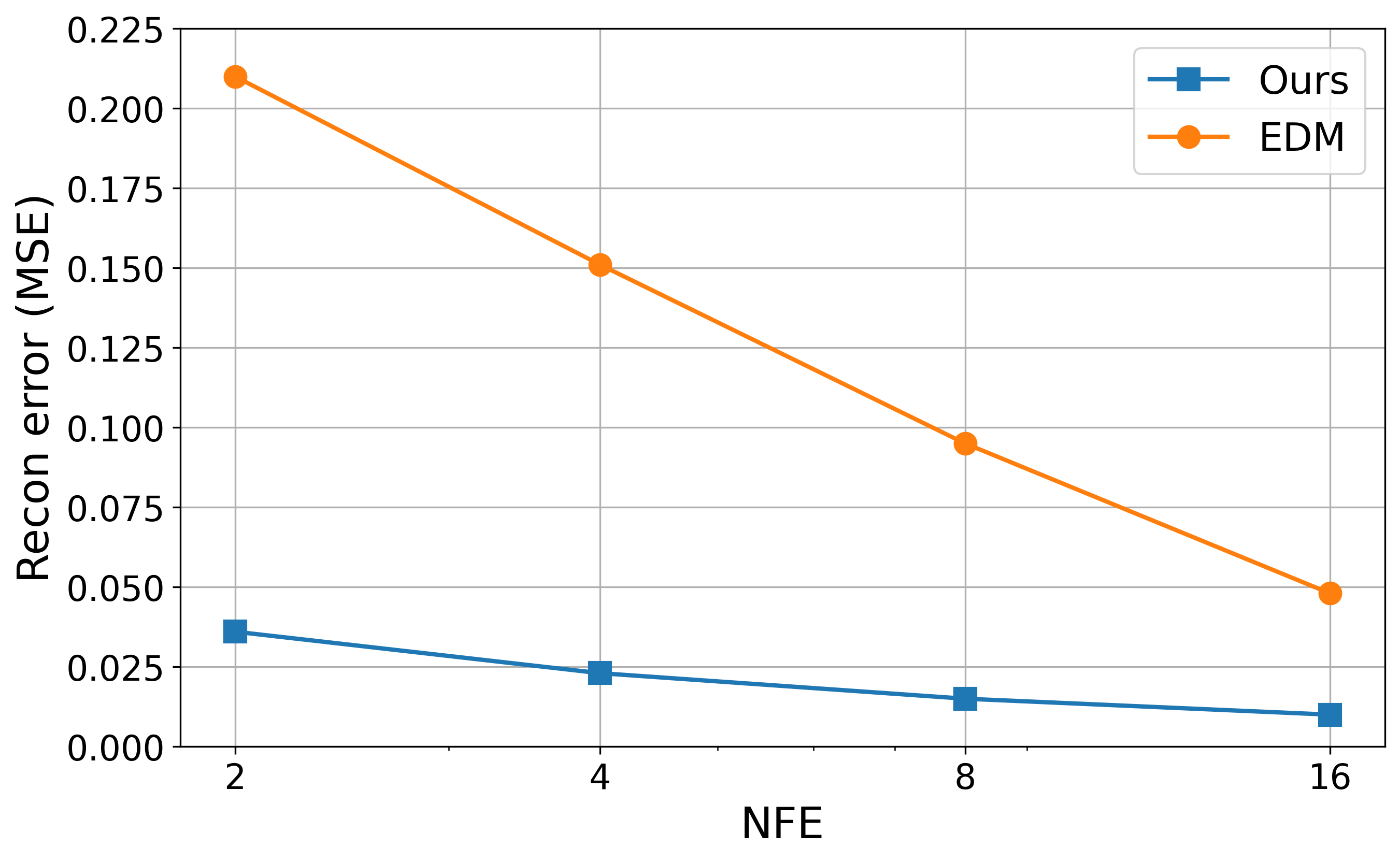}
        \caption{Reconstruction loss}
    \end{subfigure}
    \begin{subfigure}[t]{0.32\linewidth}
        \centering
        \includegraphics[height=0.65\textwidth]{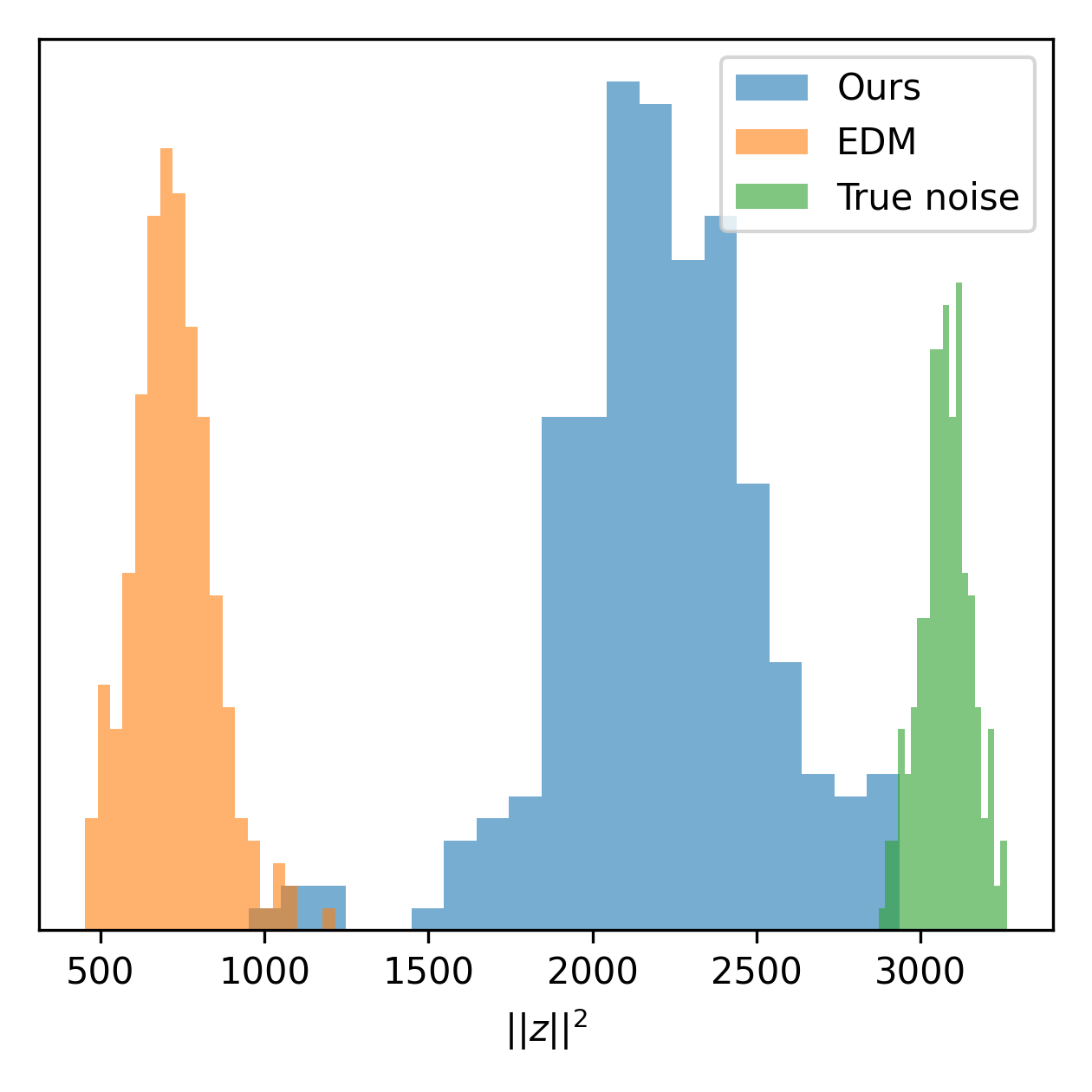}
        \caption{Distribution of $\|\mathbf{z}\|_2^2$}
    \end{subfigure}
    \begin{subfigure}[t]{0.32\linewidth}
        \centering
        \includegraphics[height=0.65\textwidth]{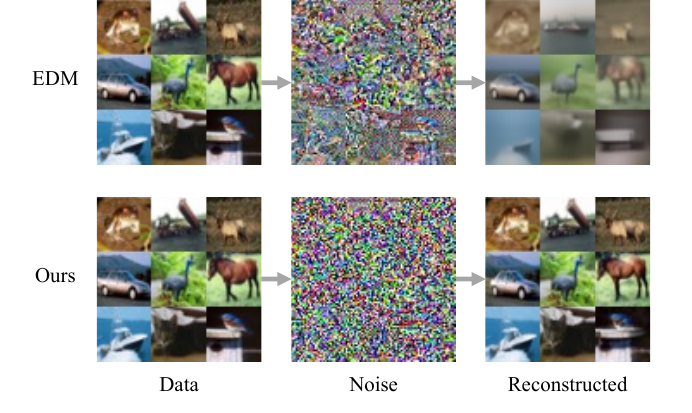}
        \caption{Inversion/reconstruction results}
    \end{subfigure}
    \caption{Inversion results on CIFAR-10. (a) Reconstruction error between real and reconstructed data is measured by the mean squared error (MSE), where the x-axis represents NFEs used for inversion and reconstruction (e.g. 2 means 2 for inversion and 2 for reconstruction).
    (b) Distribution of $||\z||_2^2$ of the inverted noises as a proxy for Gaussianity (NFE = 8). The green histogram represents the distribution of true noise, which is Chi-squared with $3 \times 32 \times 32 = 3072$ degrees of freedom.
    (c) Inversion and reconstruction results using (8 + 8) NFEs. With only 8 NFEs, EDM fails to produce realistic noise, and also the reconstructed samples are blurry.}
    \label{fig:inversion-quan}
\end{figure}

\begin{figure}[]
    \centering
    \includegraphics[width=0.95\linewidth]{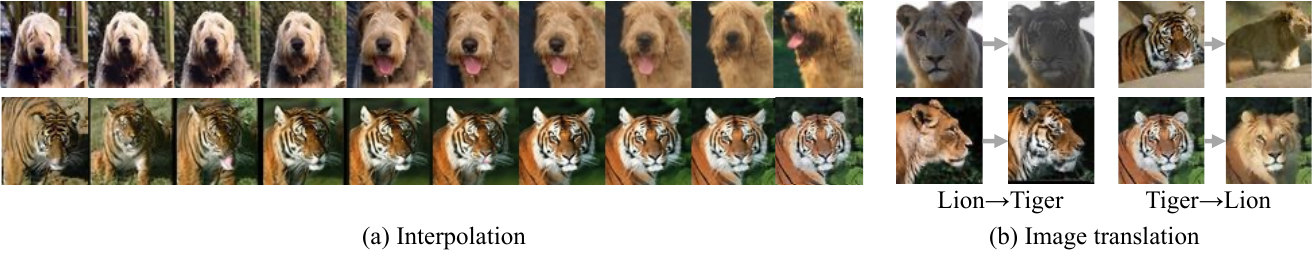}
    \caption{Applications of few-step inversion. (a) Interpolation between two real images. (b) Image-to-image translation. The total NFEs used are 6 (4 for inversion and 2 for generation).}
    \label{fig:inversion-qual}
\end{figure}

\subsection{Inversion}
Unlike distillation methods, rectified flows are neural ODEs, thus they allow for \textit{inversion} from data to noise by simply integrating the ODE in the backward direction.
In diffusion models, inversion has been used for various applications such as image editing~\citep{hertz2022prompt,kim2022diffusionclip,wallace2023edict,su2022dual,hong2023exact} and watermarking~\citep{wen2023tree},
but it usually requires many NFEs.
Figure~\ref{fig:inversion-quan} (a) demonstrates that our 2-rectified flow++ achieves significantly lower reconstruction error than EDM.
Notably, the reconstruction error of 2-rectified flow++ with only 2 NFEs is lower than that of EDM with 16 NFEs.
In (b), we compare the quality of the inverted noise, where we find that the noise vectors of 2-rectified flow are more Gaussian-like than those of EDM, in the sense that their norm is closer to that of typical Gaussian noise.
These are also shown visually in (c).
In Figure~\ref{fig:inversion-qual}, we show two applications of inversion: interpolating between two real images (a) and image-to-image translation (b).
Notably, the total NFE used is only 6 (4 for inversion and 2 for generation), which is significantly lower than what is typically required in diffusion models ($\geq 100$)~\citep{hong2023exact}.

\section{Conclusion}
In this work, we propose several improved training techniques for rectified flows, including the U-shaped timestep distribution and LPIPS-Huber loss.
We show that by combining these improvements, 2-rectified flows++ outperforms the state-of-the-art distillation methods in the 1-2 NFE regime on CIFAR-10 and ImageNet $64 \times 64$ and closes the gap with iCT, the state-of-the-art consistency model.
2-rectified flows++ have limitations though---they still do not  outperform the best consistency models (iCT), and their training is slower (by about 15\% per iteration on ImageNet) than previous rectified flows because of the LPIPS loss. Despite these shortcomings, the training techniques we propose can easily and significantly boost the performance of rectified flows in the low NFE setting, without harming performance at the higher NFE setting. 

\section*{Acknowledgments}
This work was made possible in part by the National Science Foundation under grant CCF-2338772, CNS-2325477, as well as generous support from Google, the Sloan Foundation, Intel, and Bosch. 
This work used Bridges-2 GPU at the Pittsburgh Supercomputing Center through allocation CIS240037 from the Advanced Cyberinfrastructure Coordination Ecosystem: Services \& Support (ACCESS) program, which is supported by National Science Foundation grants 2138259, 2138286, 2138307, 2137603, and 2138296 \cite{boerner2023access}.

\bibliography{reference}
\bibliographystyle{plainnat}

\newpage

\appendix


\section{Equivalence of $\vb$-parameterization and $\x$-parameterization}
\label{sec:equivalence}
Given $\x_t = (1-t)\x + t\z$ and $\z - \x = (\x_t - \x)/t$, the equivalence of the $\x$-parameterization (Eq.~\eqref{eq:objective-x}) and the $\vb$-parameterization (Eq.~\eqref{eq:objective-v}) can be shown in the following way.
The result is borrowed from the Appendix of \citet{lee2023minimizing}, and we provide here for completeness.
\begin{align}
    \int_0^1 \mathbb E[|| (\z - \x) - \vb_\theta(\x_t, t) ||_2^2]\ dt
    &= \int_0^1 \mathbb E[|| (\x_t - \x)/t - \vb_\theta(\x_t, t) ||_2^2]\ dt
    \\
    &= \int_0^1 \mathbb E[|| (\x_t - \x)/t - (\x_t - \x_\theta(\x_t, t))/t ||_2^2]\ dt
    \\
    &= \int_0^1 \mathbb E[\frac{1}{t^2}|| \x - \x_\theta(\x_t, t)) ||_2^2]\ dt.
    \label{appendix:eq:fm-loss}
\end{align}
This is equivalent to Eq.~\eqref{eq:objective-x} with $\omega(t) = 1/t^2$.

\section{Additional Details for Figure~\ref{fig:claim}}
\begin{figure}[]
    \centering
    \begin{subfigure}[t]{0.45\linewidth}
        \centering
        \includegraphics[height=0.65\textwidth]{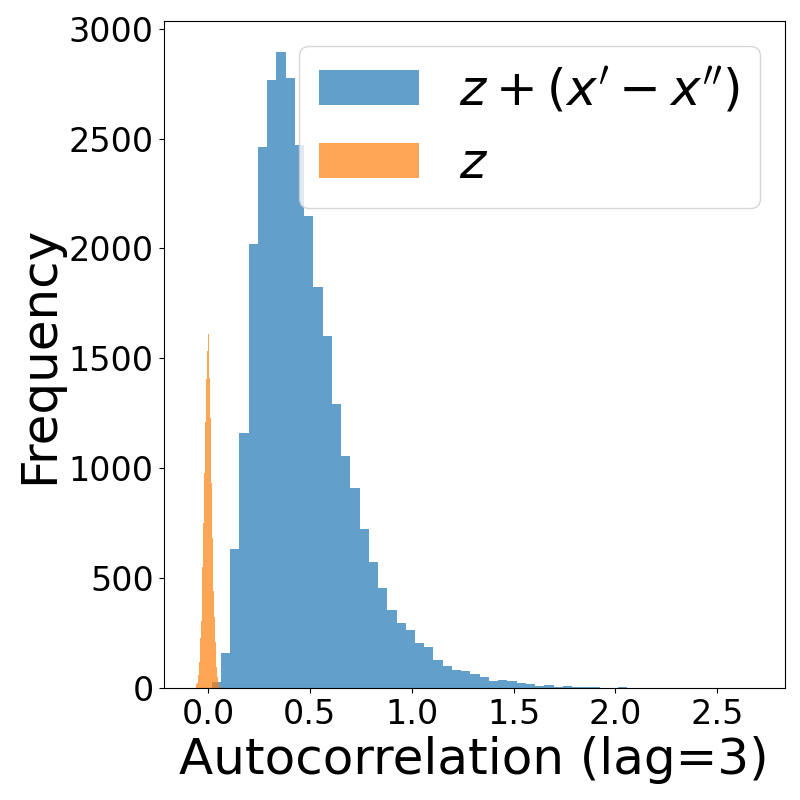}
        \caption{Reconstruction loss}
    \end{subfigure}
    \begin{subfigure}[t]{0.45\linewidth}
        \centering
        \includegraphics[height=0.65\textwidth]{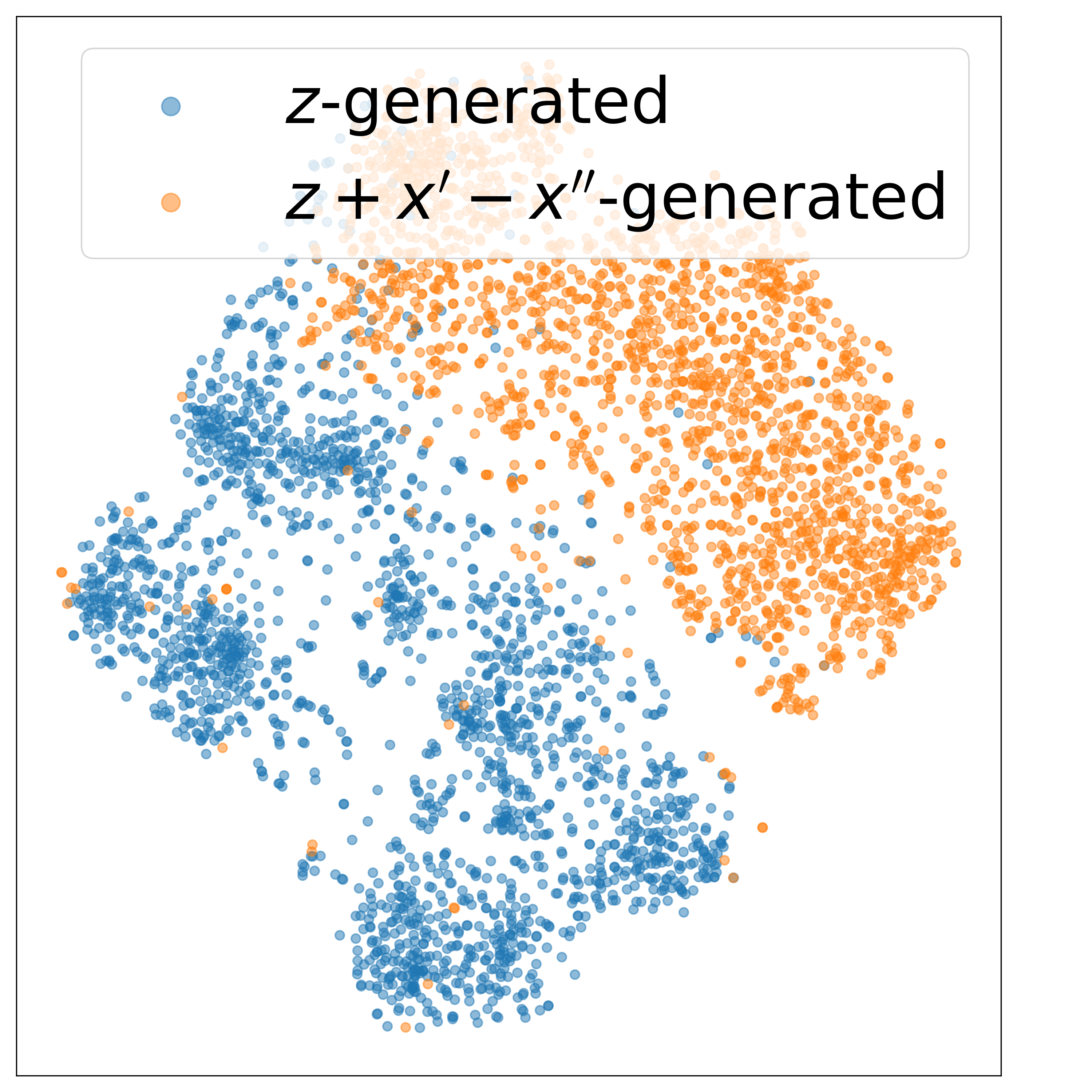}
        \caption{Distribution of $\|\mathbf{z}\|_2^2$}
    \end{subfigure}
   
    \caption{(a) Autocorrelation plot on CIFAR-10 analogous to Figure~\ref{fig:claim}(d). (b) t-SNE visualization of the inception-v3 features of the samples in Figure~\ref{fig:claim}(b). They show negligible overlap.}
    \label{fig:claim-appendix}
\end{figure}

In this section, we provide additional results details for Figure~\ref{fig:claim}. Figure~\ref{fig:claim-appendix}(a) shows the autocorrelation histogram on CIFAR-10 while Figure~\ref{fig:claim}(d) is on FFHQ-64.
Figure~\ref{fig:claim-appendix}(b) shows that the inception features of the samples in Figure~\ref{fig:claim}(b) rarely overlap with each other.

For the autocorrelation plots, we use 30,000 pairs of $(\x', \x'')$ and randomly sample $\z$ from the standard Gaussian distribution. For a $d$-dimensional vector $\ub$, we define the autocorrelation as:
\begin{align}
    R_{\ub}(l) = \frac{1}{d-l} \sum_{k=1}^{d-l} \ub_k  \ub_{k+l},
\end{align}
where $\ub_k$ is the $k$-th element of a vector $\ub$ and $l > 0$ represents the lag.

\section{Initialization Results}
\label{app:initialization_results}
\subsection{Proof of Proposition~\ref{prop:conversion}}
\label{sec:proof-prop}


Consider two perturbation kernels $p_t(\x_t | \x) = \mathcal N(s(t) \x, \sigma(t)^2 \Ib)$ and $p_t'(\x_t | \x) = \mathcal N(s'(t) \x, \sigma'(t)^2 \Ib)$:
\begin{align}
    p_t(\x_t | \x) = \frac{1}{(2\pi \sigma(t)^2)^{d/2}} \exp(-\frac{1}{2\sigma(t)^2} ||\x_t - s(t) \x||_2^2) \\
    p_{t}'(\x_{t} | \x) = \frac{1}{(2\pi \sigma'(t)^2)^{d/2}} \exp(-\frac{1}{2\sigma'(t)^2} ||\x_{t} - s'(t) \x||_2^2).
\end{align}
Let $t'(t)$ be such that $\frac{s(t)}{\sigma(t)} = \frac{s'(t')}{\sigma'(t')}$.
We will show that $p_t(\x | \x_t) = p_{t'}'(\x | \frac{s'(t')}{s(t)}\x_t)$.
We start by showing:
\begin{align}
    p_{t'}'(\frac{s'(t')}{s(t)}\x_{t} | \x) &= \frac{1}{(2\pi \sigma'(t')^2)^{d/2}} \exp(-\frac{1}{2\sigma'(t')^2} ||\frac{s'(t')}{s(t)}\x_{t} - s'(t') \x||_2^2) \\
    &= \frac{1}{(2\pi \sigma'(t')^2)^{d/2}} \exp(-\frac{1}{2\sigma'(t')^2} ||\frac{s'(t')}{s(t)} (\x_{t} - s(t) \x)||_2^2) \\
    &= \frac{1}{(2\pi \sigma'(t')^2)^{d/2}} \exp(-\frac{1}{2\sigma'(t')^2} \frac{s'(t')^2}{s(t)^2}|| \x_{t} - s(t) \x||_2^2) \\
    &= \frac{1}{(2\pi \sigma'(t')^2)^{d/2}} \exp(-\frac{1}{2\sigma(t)^2} \frac{s(t)^2}{s(t)^2}|| \x_{t} - s(t) \x||_2^2) \\
    &= \frac{1}{(2\pi \sigma'(t')^2)^{d/2}} \exp(-\frac{1}{2\sigma(t)^2} || \x_{t} - s(t) \x||_2^2) \\
    &= \frac{1}{(2\pi \sigma'(t')^2)^{d/2}} \frac{(2\pi \sigma(t)^2)^{d/2}}{(2\pi \sigma(t)^2)^{d/2}}  \exp(-\frac{1}{2\sigma(t)^2} || \x_{t} - s(t) \x||_2^2) \\
    &= \frac{(2\pi \sigma(t)^2)^{d/2}}{(2\pi \sigma'(t')^2)^{d/2}} p_t(\x_t | \x) \\
    &= \left (\frac{\sigma(t)}{\sigma '(t')}\right )^{d} p_t(\x_t | \x). \label{eq:posterior-propto}
\end{align} 
Here, Eq.~\eqref{eq:posterior-propto} says that $p_t(\x_t | \x) \propto p_{t'}'(\frac{s'(t')}{s(t)}\x_t | \x)$ (but not equal), which is a minor fix from the original proof~\citep{pokle2023training}.
Then, we have
\begin{align}
    & p_t(\x | \x_t) = \frac{1}{p_t(\x_t)} p_\x (\x) p_t(\x_t | \x) \\
    & p_{t'}'(\x | \frac{s'(t')}{s(t)}\x_t) = \left (\frac{\sigma(t)}{\sigma '(t')}\right )^{d} \frac{1}{p_{t'}'(\frac{s'(t')}{s(t)}\x_t)} p_\x (\x)  p_t(\x_t | \x) 
\end{align}
Since $p_{t'}'(\x | \frac{s'(t')}{s(t)}\x_t)$ should be integrated to one, $\left (\frac{\sigma(t)}{\sigma '(t')}\right )^{d} \frac{1}{p_{t'}'(\frac{s'(t')}{s(t)}\x_t)} = \int p_\x (\x)  p_t(\x_t | \x) d\x$ and thus two densities are equal.
As the posterior densities are the same, their expectations are also the same.

In our case, $s(t) = 1-t$, $\sigma(t) = t$, and $p_t' (\x_t | \x)$ is the perturbation kernel of either the VP or VE diffusion model.
Now we have to find $t'$ such that $\frac{1-t}{t} = \frac{s'(t')}{\sigma'(t')}$.

\subsection{Perturbation Kernel Instantiations}
We next provide the values of the converted time and scale of the variance preserving (VP) and variance exploding (VE) diffusion models.

\paragraph{VE diffusion model}
\citet{karras2022elucidating} defines the perturbation kernel of the VE diffusion model as $p'_t(\x_t | \x) = \mathcal N(\x, t^2 \Ib)$.
Then, $t'$ satisfies $\frac{1-t}{t} = \frac{s'(t')}{\sigma'(t')} = \frac{1}{t'}$, so $t' = \frac{t}{1-t}$, and $\frac{s'(t')}{s(t)} = \frac{1}{1-t}$, which correspond to $t_{\text{VE}}$ and $s_{\text{VE}}$ in Table~\ref{tab:time-scale}.

\paragraph{VP diffusion model}
\citet{song2020score} defines the perturbation kernel of the VP diffusion model as $p'_t(\x_t | \x) = \mathcal N(\alpha(t) \x, (1-\alpha(t)^2) \Ib)$, where $\alpha(t) := \exp(-\frac{1}{2} \int_0^t (19.9s + 0.1) ds)$ defined on $t \in [0, 1]$.
Then, $t'$ satisfies $\frac{1-t}{t} = \frac{s'(t')}{\sigma'(t')} = \frac{\alpha(t')}{\sqrt{1-\alpha(t')^2}}$.
From here, we have
\begin{align}
    \frac{(1-t)^2}{t^2} = \frac{\alpha(t')^2}{1-\alpha(t')^2} \\
    \alpha(t') = \sqrt{\frac{(1-t)^2}{t^2 + (1-t)^2}},
\end{align}
where we used the fact that $\alpha(t) > 0$.
Since $\alpha(t)$ is a monotonically decreasing function for $t \geq 0$, we can use its inverse $\alpha^{-1}$ to find $t' = \alpha^{-1}(\sqrt{\frac{(1-t)^2}{t^2 + (1-t)^2}})$.

\begin{align}
    &y = \alpha(t) = \exp(-\frac{1}{2} \int_0^t (19.9s + 0.1) ds) = \exp(-\frac{19.9}{4}t^2 - 0.05t) \\
    &\ln y = -\frac{19.9}{4}t^2 - 0.05t \\
    &\frac{19.9}{4}t^2 + 0.05t + \ln y = 0
\end{align}

Applying the quadratic formula, we have
\begin{align}
    t = \frac{-0.05 \pm \sqrt{0.05^2 - 4 \cdot \frac{19.9}{4} \ln y}}{2 \cdot \frac{19.9}{4}} = \frac{-0.05 \pm \sqrt{0.0025 - 19.9 \ln y}}{9.95}.
\end{align}
Since $y=\alpha(t)$ is monotonically decreasing, we can choose the positive root:
\begin{align}
    \alpha^{-1}(y) = \frac{-0.05 + \sqrt{0.0025 - 19.9 \ln y}}{9.95}.
\end{align}

Now, we arrive at
\begin{align}
    t' = \alpha^{-1}(\sqrt{\frac{(1-t)^2}{t^2 + (1-t)^2}}) = \frac{-0.05 + \sqrt{0.0025 - 19.9 \ln \sqrt{\frac{(1-t)^2}{t^2 + (1-t)^2}}}}{9.95},
\end{align}
which corresponds to $t_{\text{VP}}$ in Table~\ref{tab:time-scale}. Also, we have $\frac{s'(t')}{s(t)} = \frac{\alpha(t')}{1-t}$, which is $s_{\text{VP}}$ in Table~\ref{tab:time-scale}.

\section{New Update Rule}
\label{sec:pseudo-code}
In the standard Euler solver, the update rule is $\z_{t-\Delta t} := \z_t - \vb(\z_t, t) \Delta t$.
Alternatively, as $\x_{\thetab} (\z_t, t)$ of our model generates pretty good samples, we can instead use the linear interpolation between $\x_{\thetab} (\z_t, t)$ and $\z_1$ to get the next step:
$\z_{t-\Delta t} := (1-(t - \Delta t)) \x_{\thetab} (\z_t, t) + (t - \Delta t) \z_1$.
Note that when NFE $< 3$, the two update rules are equivalent and do not affect our results in Section \ref{sec:unconditional}.
Fig.~\ref{fig:multi} shows that when applied to existing solvers, the new update rule improves the sampling efficiency up to $4\times$, achieving the best FID with $\leq 5$ NFEs.

Algorithm~\ref{alg:generate} shows the pseudocode for generating samples using the new update rule.
Unlike the standard Euler update rule which only depends on the current state $\z_t$, our new update rule utilizes the previous state (i.e., $\z_1$) to generate the next state $\z_{t-\Delta t}$ and thus can be viewed as a form of history-dependent samplers.
Obviously, incorporating the initial state only would not be the best choice. We believe that the result can be further improved, especially by using learning-based solvers~\citep{watson2021learning,shaul2023bespoke}; and leave such exploration to future work.

\begin{algorithm}[t]
    \caption{Generate}
    \label{alg:generate}
    \definecolor{codeblue}{rgb}{0.25,0.5,0.5}
    \lstset{
      backgroundcolor=\color{white},
      basicstyle=\fontsize{7.2pt}{7.2pt}\ttfamily\selectfont,
      columns=fullflexible,
      breaklines=true,
      captionpos=b,
      commentstyle=\fontsize{7.2pt}{7.2pt}\color{codeblue},
      keywordstyle=\fontsize{7.2pt}{7.2pt},
    }
    \begin{lstlisting}[language=python]
    def generate(z1, label, model, time_schedule, N, solver, sampler, device):
        """
        z1: initial noise
        label: class label
        model: v_theta
        time_schedule: time schedule, e.g., [0.99999, 0.5, 0] for 2 steps
        N: NFE
        solver: 'euler' or 'heun'
        sampler: 'default' or 'new'
        """
        
        z = z1.clone()
        cnt = 0
        for i in range(len(time_schedule[:-1])):
            t = torch.ones((z.shape[0]), device=device) * time_schedule[i]
            t_next = torch.ones((z.shape[0]), device=device) * time_schedule[i+1]
            dt = t_next[0] - t[0]
            vt = model(z, t, label)
            x0hat = z - vt * t.view(-1,1,1,1)
            if solver == 'heun' and cnt < N - 1: # Heun correction
                if sampler == 'default':
                    z_next = z.detach().clone() + vt * dt
                elif sampler == 'new':
                    z_next = (1 - t_next.view(-1,1,1,1)) * x0hat + t_next.view(-1,1,1,1) * z1
                vt_next = model(z_next, t_next, label)
                vt = (vt + vt_next) / 2
                x0hat = z - vt * t.view(-1,1,1,1)
            if sampler == 'default':
                z = z.detach().clone() + vt * dt
            elif sampler == 'new':
                z = (1 - t_next.view(-1,1,1,1)) * x0hat + t_next.view(-1,1,1,1) * z1
            cnt += 1
            
        return z
    \end{lstlisting}
    \end{algorithm}
    
\section{Experimental Details}
\label{sec:exp-details}
Before training 2-rectified flow, we generate data-noise pairs following the sampling regime of EDM~\citep{karras2022elucidating}.
For CIFAR-10, we generate 1M pairs using 35 NFEs.
For AFHQ, FFHQ, and ImageNet, we generate 5M pairs using 79 NFEs.
We use Heun's second-order solver for all cases.
In Table~\ref{tab:cifar-10}, we report the result of config G in Table~\ref{tab:ablation}.
In ImageNet, we use the batch size of 2048 and train the models for 700,000 iterations using mixed-precision training~\citep{micikevicius2017mixed} with the dynamic loss scaling.
We use config E in Table~\ref{tab:ablation} for ImageNet.

We provide training configurations in Table~\ref{tab:config}. For all datasets, we use Adam optimizer. We use the exponential moving average (EMA) with 0.9999 decay rate for all datasets.

On ImageNet, the training takes roughly 9 days with 64 NVIDIA V100 GPUs.
On CIFAR-10 and FFHQ/AFHQ, it takes roughly 4 days with 16 and 8 V100 GPUs, respectively.
For all cases, we use the NVIDIA DGX-2 cluster.
To prevent zero-division error with EDM initialization, we sample $t$ from $[0.00001, 0.99999]$ in practice.
For a two-step generation, we evaluate $\vb_{\thetab}$ at $t=0.99999$ and $t=0.8$. For other NFEs, we uniformly divide the interval $[0.00001, 0.99999]$.

\begin{table}[]
\centering
\caption{Training configurations for each dataset. We linearly ramp up learning rates for all datasets.}
\begin{tabular}{@{}lcccc@{}}
\toprule
Datasets    & Batch size & Dropout & Learning rate & Warm up iter. \\ \midrule
CIFAR-10    & 512        & 0.13    & 2e-4          & 5000          \\
FFHQ / AFHQ & 256        & 0.25    & 2e-4          & 5000          \\
ImageNet    & 2048       & 0.10    & 1e-4          & 2500          \\ \bottomrule
\end{tabular}
\label{tab:config}
\end{table}

\paragraph{License}
The following are licenses for each dataset we use:
\begin{itemize}
    \item CIFAR-10: Unknown
    \item FFHQ: CC BY-NC-SA 4.0
    \item AFHQ: CC BY-NC 4.0
    \item ImageNet: Custom (research, non-commercial)
\end{itemize}

\section{Broader Impacts}
\label{sec:impact}
This paper proposes an advanced algorithm to generate realistic data at high speed, which could have both positive and negative impacts. For example, it could be used for generating malicious or misleading content. Therefore, such technology should be deployed and used responsibly and with caution. We believe that our work is not expected to have any more potential negative impact than other work in the field of generative modeling.

\newpage    
\section{Uncurated Synthetic Samples}
\label{sec:sample}
\begin{figure}[t]
    \centering
    \includegraphics[width=1.\linewidth]{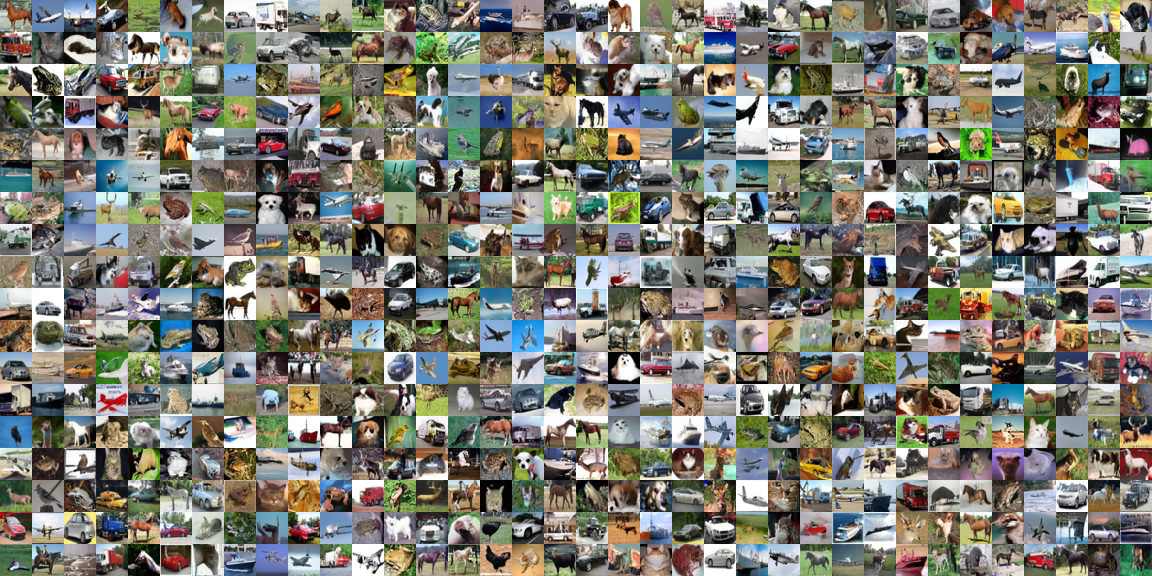}
    \vspace{-2mm}
    \caption{Synthetic samples from 2-rectified flow++ on CIFAR-10 with NFE = 1 (FID=3.38).}
    \label{fig:sample-cifar-1}
\end{figure}

\begin{figure}[b]
    \centering
    \includegraphics[width=1.\linewidth]{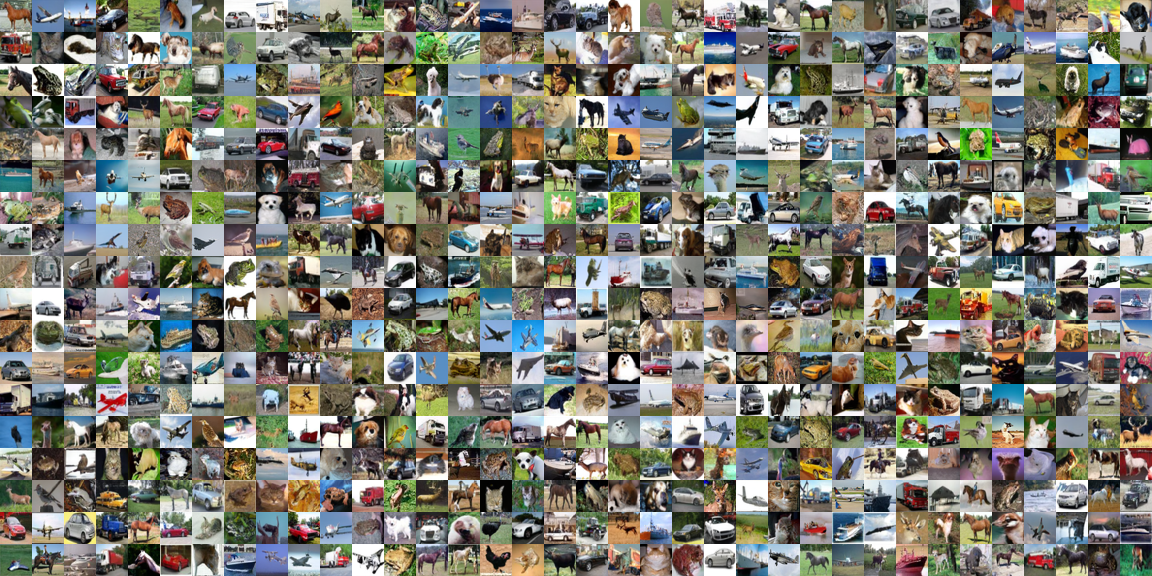}
    \vspace{-2mm}
    \caption{Synthetic samples from 2-rectified flow++ on CIFAR-10 with NFE = 2 (FID=2.76).}
    \label{fig:sample-cifar-2}
\end{figure}

\begin{figure}[]
    \centering
    \includegraphics[width=1.\linewidth]{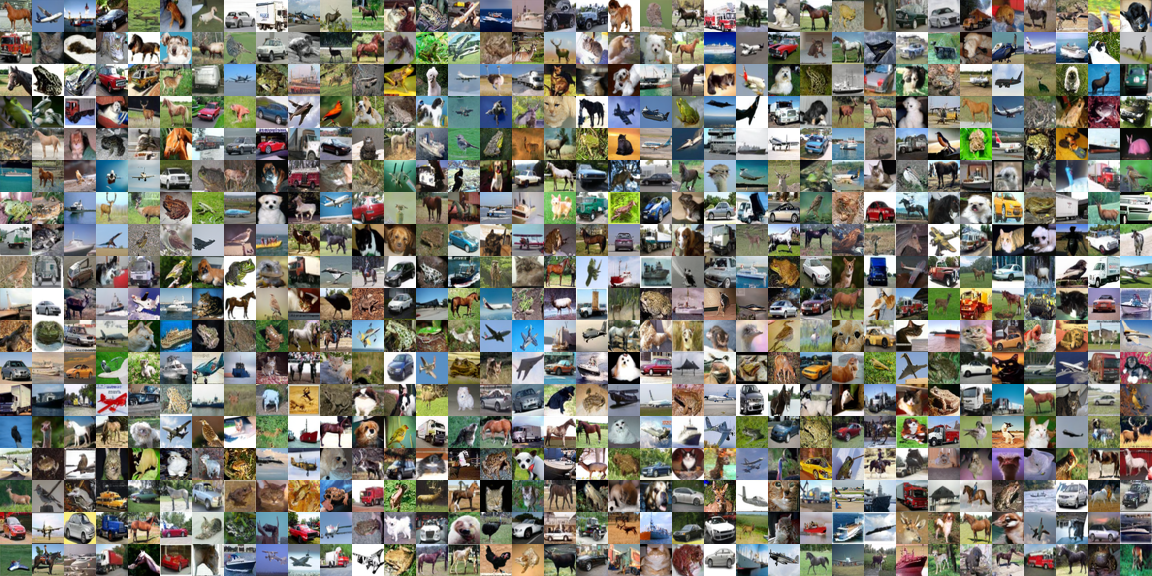}
    \vspace{-2mm}
    \caption{Synthetic samples from 2-rectified flow++ on CIFAR-10 with NFE = 4 (FID=2.50).}
    \label{fig:sample-cifar-4}
\end{figure}

\begin{figure}[]
    \centering
    \includegraphics[width=1.\linewidth]{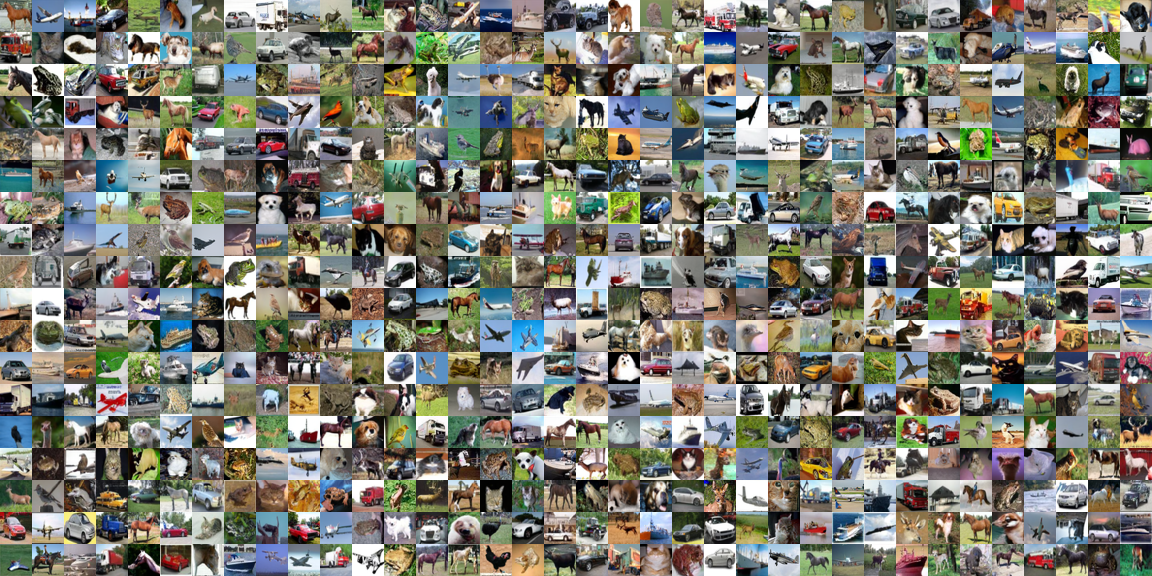}
    \vspace{-2mm}
    \caption{Synthetic samples from 2-rectified flow++ on CIFAR-10 with NFE = 5 (FID=2.45).}
    \label{fig:sample-cifar-5}
\end{figure}

\begin{figure}[]
    \centering
    \includegraphics[width=1.\linewidth]{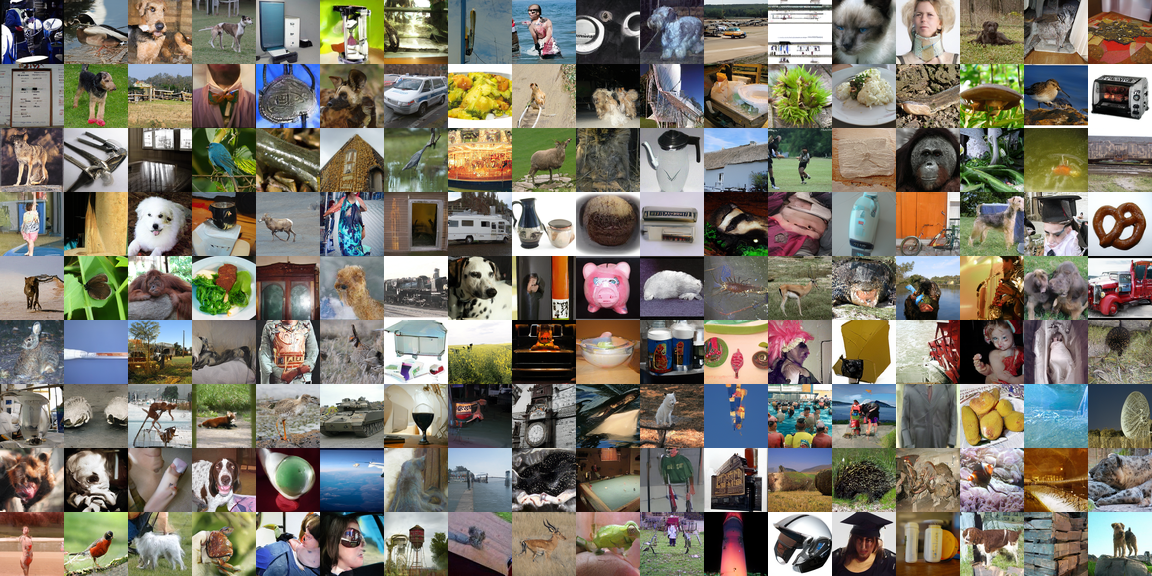}
    \vspace{-2mm}
    \caption{Synthetic samples from 2-rectified flow++ on ImageNet 64$\times$64 with NFE = 1 (FID=4.31).}
    \label{fig:sample-imagenet-1}
\end{figure}

\begin{figure}[]
    \centering
    \includegraphics[width=1.\linewidth]{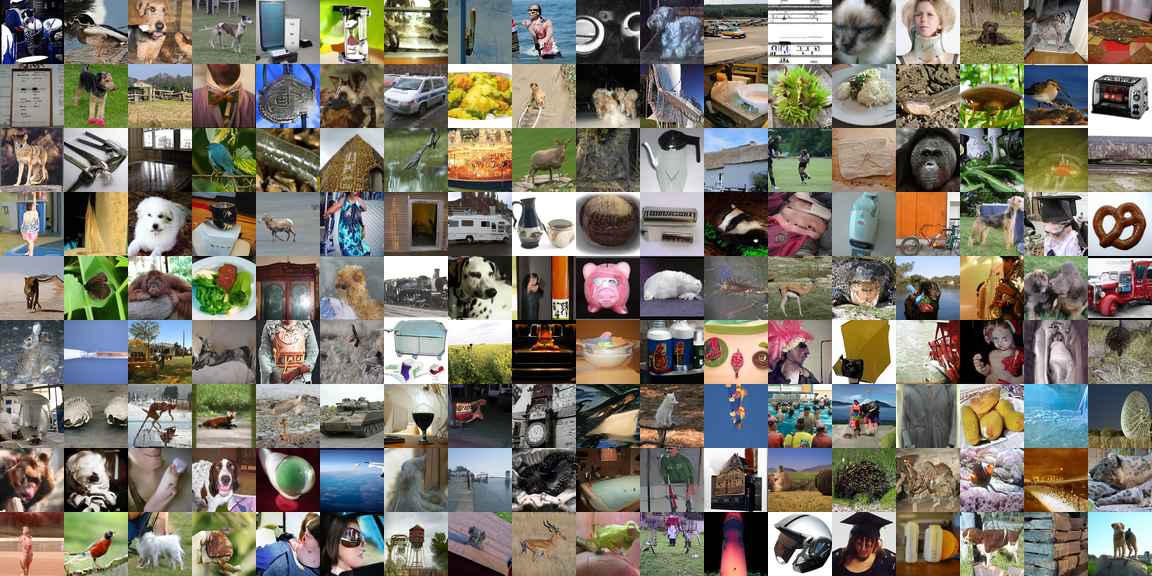}
    \vspace{-2mm}
    \caption{Synthetic samples from 2-rectified flow++ on ImageNet 64$\times$64 with NFE = 2 (FID=3.64).}
    \label{fig:sample-imagenet-2}
\end{figure}

\begin{figure}[]
    \centering
    \includegraphics[width=1.\linewidth]{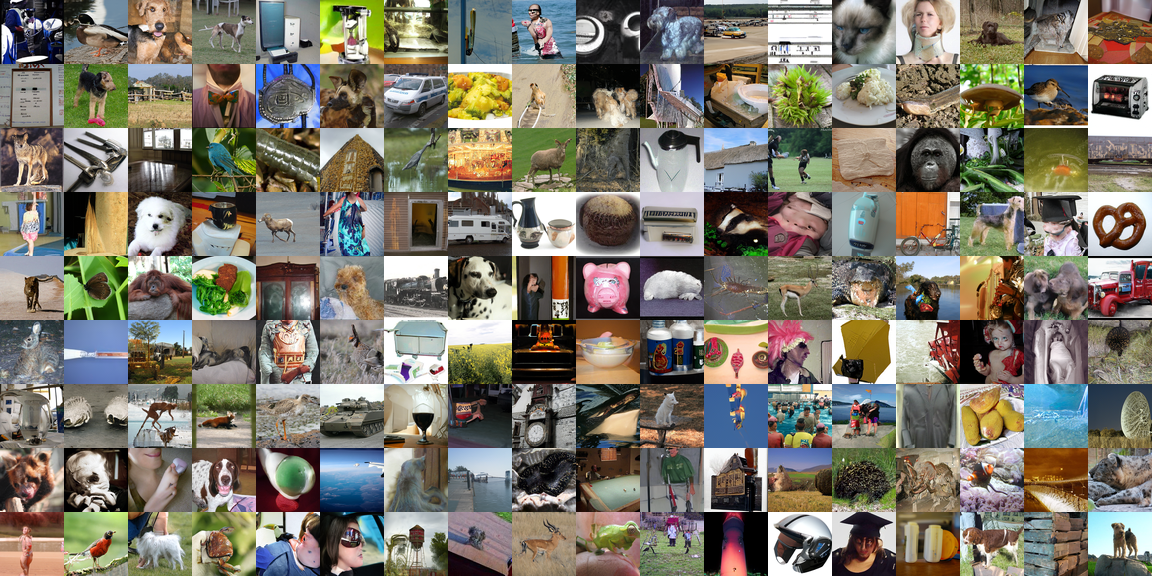}
    \vspace{-2mm}
    \caption{Synthetic samples from 2-rectified flow++ on ImageNet 64$\times$64 with NFE = 4 (FID=3.44).}
    \label{fig:sample-imagenet-4}
\end{figure}

\begin{figure}[]
    \centering
    \includegraphics[width=1.\linewidth]{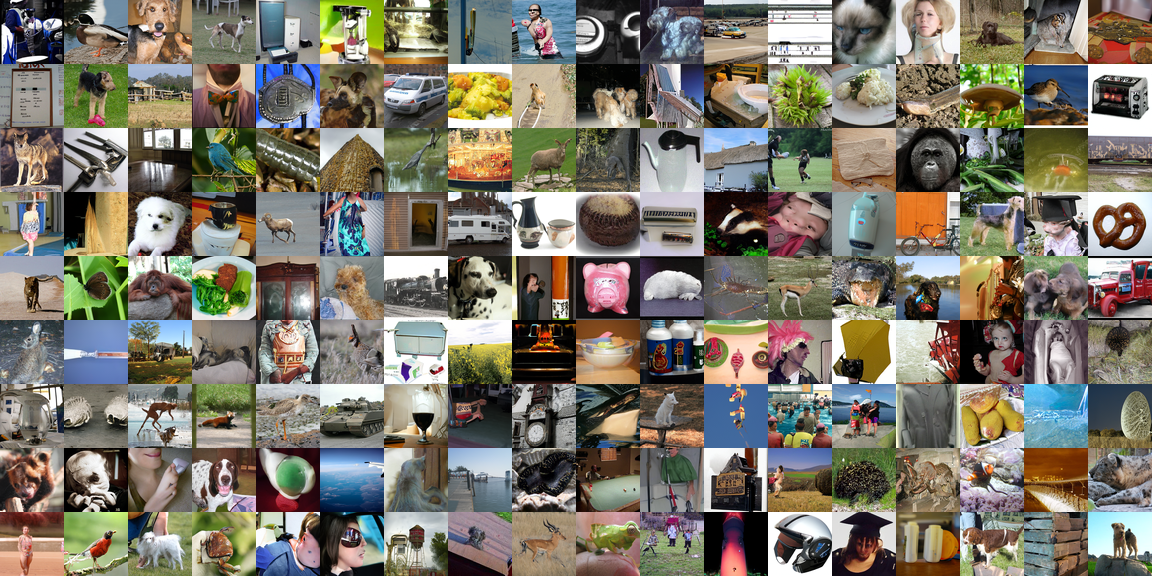}
    \vspace{-2mm}
    \caption{Synthetic samples from 2-rectified flow++ on ImageNet 64$\times$64 with NFE = 8 (FID=3.32).}
    \label{fig:sample-imagenet-8}
\end{figure}

\begin{figure}[]
    \centering
    \includegraphics[width=1.\linewidth]{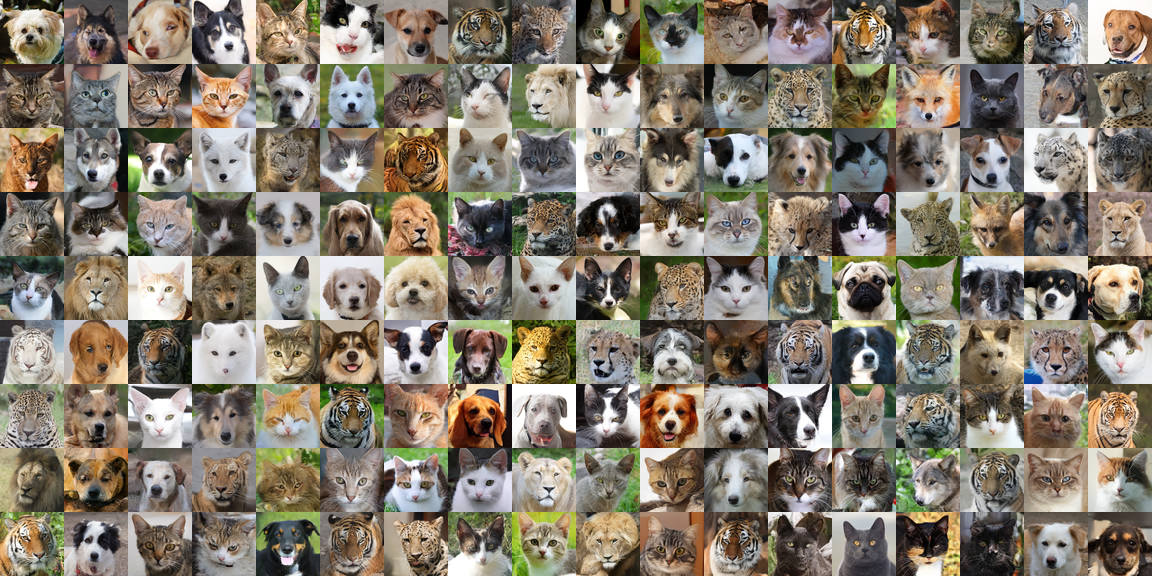}
    \vspace{-2mm}
    \caption{Synthetic samples from 2-rectified flow++ on AFHQ 64$\times$64 with NFE = 1 (FID=4.11).}
    \label{fig:sample-afhq-1}
\end{figure}

\begin{figure}[]
    \centering
    \includegraphics[width=1.\linewidth]{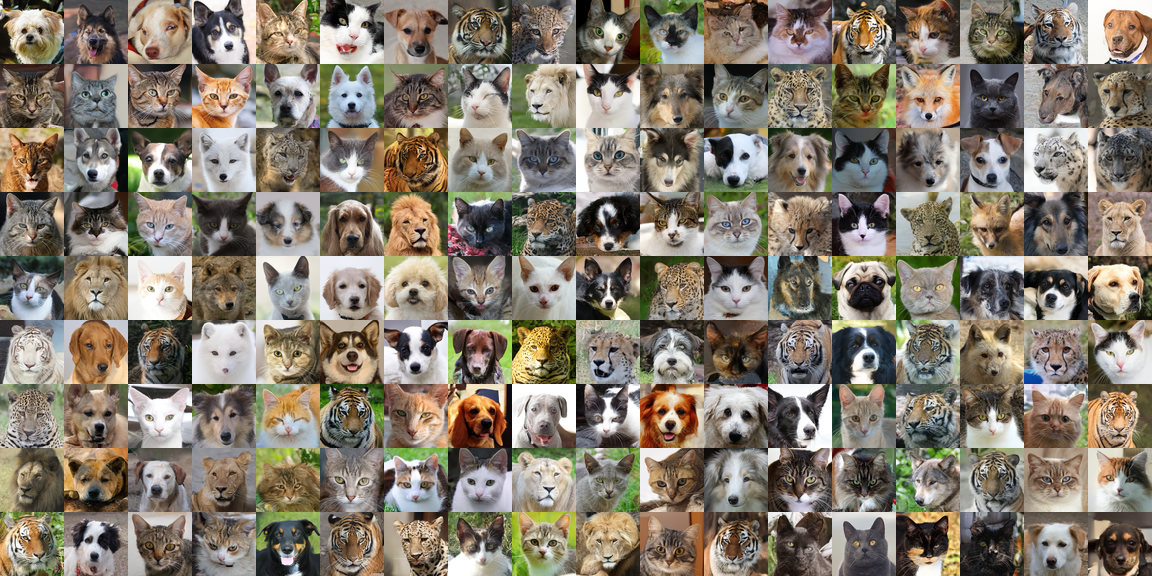}
    \vspace{-2mm}
    \caption{Synthetic samples from 2-rectified flow++ on AFHQ 64$\times$64 with NFE = 2 (FID=3.12).}
    \label{fig:sample-afhq-2}
\end{figure}

\begin{figure}[]
    \centering
    \includegraphics[width=1.\linewidth]{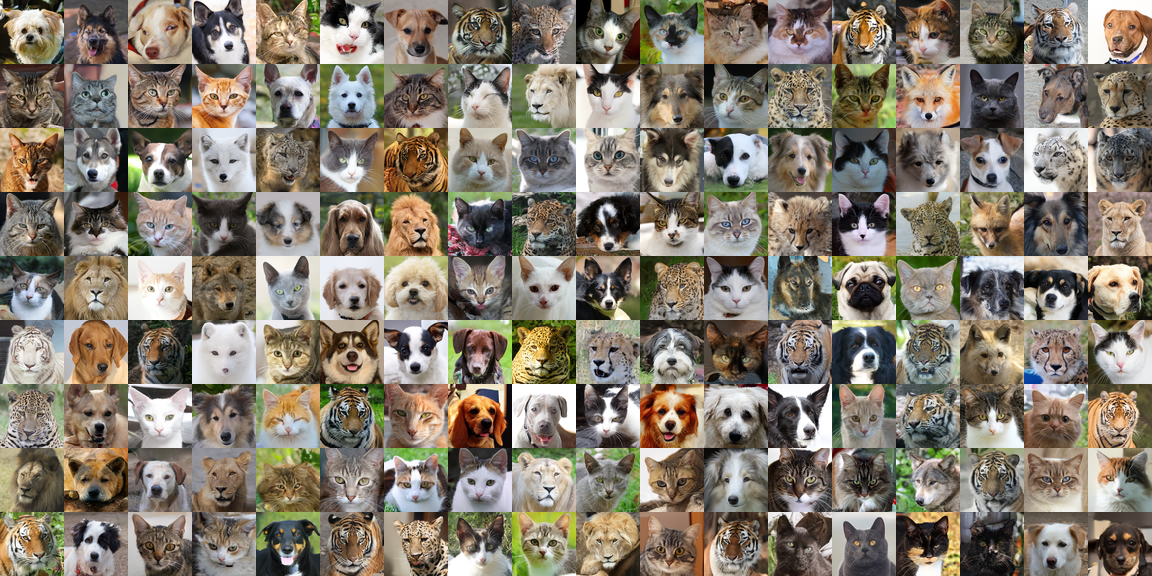}
    \vspace{-2mm}
    \caption{Synthetic samples from 2-rectified flow++ on AFHQ 64$\times$64 with NFE = 4 (FID=2.90).}
    \label{fig:sample-afhq-4}
\end{figure}

\begin{figure}[]
    \centering
    \includegraphics[width=1.\linewidth]{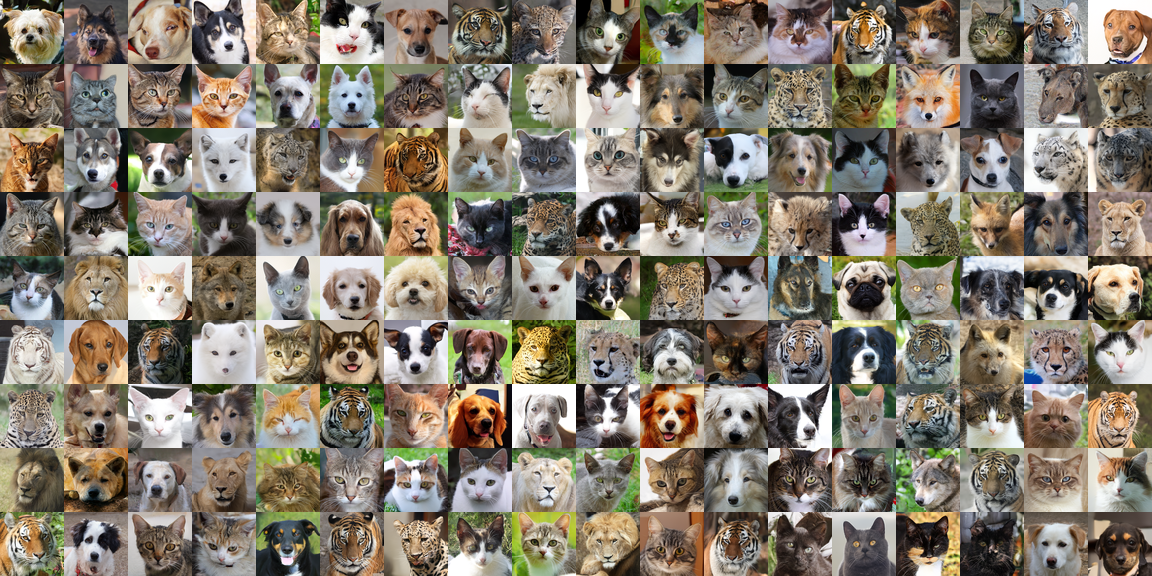}
    \vspace{-2mm}
    \caption{Synthetic samples from 2-rectified flow++ on AFHQ 64$\times$64 with NFE = 5 (FID=2.86).}
    \label{fig:sample-afhq-5}
\end{figure}

\begin{figure}[]
    \centering
    \includegraphics[width=1.\linewidth]{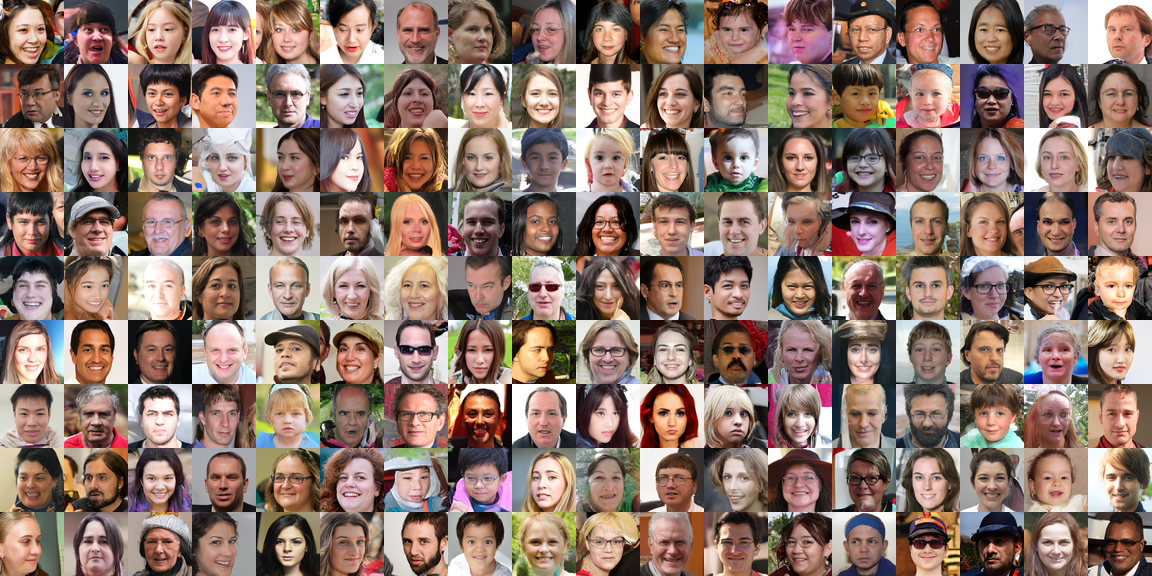}
    \vspace{-2mm}
    \caption{Synthetic samples from 2-rectified flow++ on FFHQ 64$\times$64 with NFE = 1 (FID=5.21).}
    \label{fig:sample-ffhq-1}
\end{figure}

\begin{figure}[]
    \centering
    \includegraphics[width=1.\linewidth]{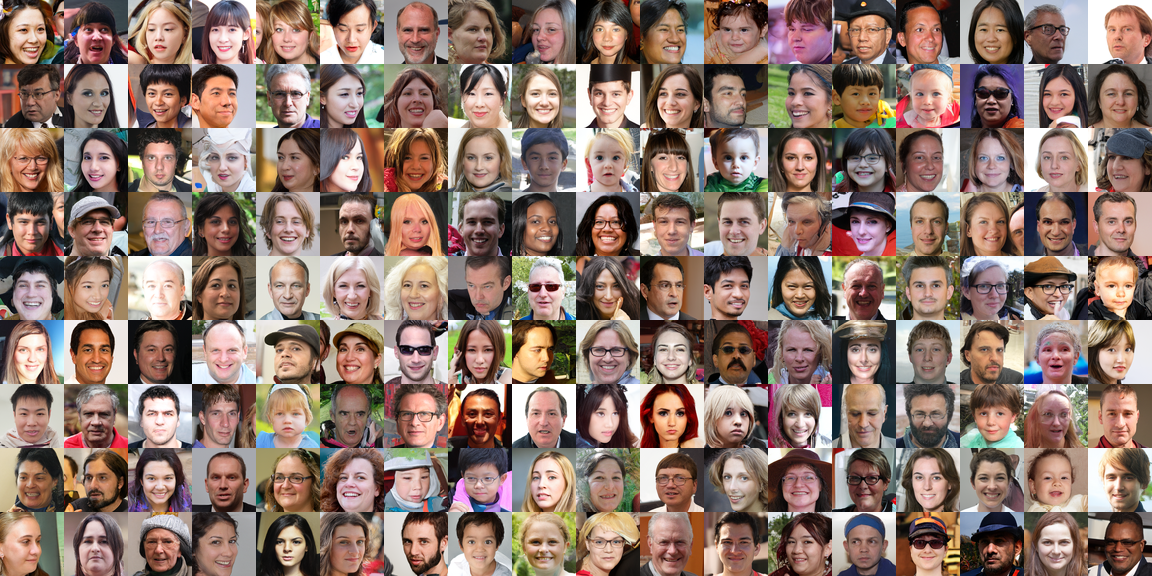}
    \vspace{-2mm}
    \caption{Synthetic samples from 2-rectified flow++ on FFHQ 64$\times$64 with NFE = 2 (FID=4.26).}
    \label{fig:sample-ffhq-2}
\end{figure}

We provide uncurated synthetic samples from our 2-rectified flow++ on CIFAR-10, AFHQ, and ImageNet in Figures~\ref{fig:sample-cifar-1}, \ref{fig:sample-cifar-2}, \ref{fig:sample-cifar-4}, \ref{fig:sample-cifar-5}, \ref{fig:sample-ffhq-1}, \ref{fig:sample-ffhq-2}, \ref{fig:sample-afhq-1}, \ref{fig:sample-afhq-2}, \ref{fig:sample-afhq-4}, \ref{fig:sample-afhq-5}, \ref{fig:sample-imagenet-1}, \ref{fig:sample-imagenet-2}, \ref{fig:sample-imagenet-4}, and \ref{fig:sample-imagenet-8}.
We use our new sampler (Sec.~\ref{sec:ode}) to generate these images.

\clearpage

\clearpage
\section*{NeurIPS Paper Checklist}

\begin{enumerate}

\item {\bf Claims}
    \item[] Question: Do the main claims made in the abstract and introduction accurately reflect the paper's contributions and scope?
    \item[] Answer: \answerYes{} 
    \item[] Justification: The abstract and introduction reflect the paper's contributions and scope such as improved empirical performance, which is backed up by our experiments in  Sec.~\ref{sec:exp}.

\item {\bf Limitations}
    \item[] Question: Does the paper discuss the limitations of the work performed by the authors?
    \item[] Answer: \answerYes{} 
    \item[] Justification: The limitations are discussed in the conclusion, such as the increased training time of our method.

\item {\bf Theory Assumptions and Proofs}
    \item[] Question: For each theoretical result, does the paper provide the full set of assumptions and a complete (and correct) proof?
    \item[] Answer: \answerYes{} 
    \item[] Justification: 
    Our only theoretical result is  Proposition~\ref{prop:conversion}, for which we provide the proof in Appendix~\ref{sec:proof-prop}. The argument in Sec.~\ref{sec:claim} is intuitive, and we do not prove it theoretically.

    \item {\bf Experimental Result Reproducibility}
    \item[] Question: Does the paper fully disclose all the information needed to reproduce the main experimental results of the paper to the extent that it affects the main claims and/or conclusions of the paper (regardless of whether the code and data are provided or not)?
    \item[] Answer: \answerYes{} 
    \item[] Justification: We provide our experimental details in Sec.~\ref{sec:exp} and App. \ref{sec:exp-details}. For the new update algorithm, we provide full pseudocode in Sec.~\ref{sec:pseudo-code}, and we will release code publicly as soon as we obtain approval to do so. 

\item {\bf Open access to data and code}
    \item[] Question: Does the paper provide open access to the data and code, with sufficient instructions to faithfully reproduce the main experimental results, as described in supplemental material?
    \item[] Answer: \answerNo{} 
    \item[] Justification: The datasets we used are publicly available. We will release the code publicly as soon as we obtain internal approval.

\item {\bf Experimental Setting/Details}
    \item[] Question: Does the paper specify all the training and test details (e.g., data splits, hyperparameters, how they were chosen, type of optimizer, etc.) necessary to understand the results?
    \item[] Answer: \answerYes{} 
    \item[] Justification: We provide these details in Sec.~\ref{sec:exp-details}.

\item {\bf Experiment Statistical Significance}
    \item[] Question: Does the paper report error bars suitably and correctly defined or other appropriate information about the statistical significance of the experiments?
    \item[] Answer: \answerYes{} 
    \item[] Justification: The shaded area in Figure~\ref{fig:loss} indicates the standard deviation. We only compute FID once due to cost constraints.

\item {\bf Experiments Compute Resources}
    \item[] Question: For each experiment, does the paper provide sufficient information on the computer resources (type of compute workers, memory, time of execution) needed to reproduce the experiments?
    \item[] Answer: \answerYes{} 
    \item[] Justification: We provide details in Sec.~\ref{sec:exp-details}.
    
\item {\bf Code Of Ethics}
    \item[] Question: Does the research conducted in the paper conform, in every respect, with the NeurIPS Code of Ethics \url{https://neurips.cc/public/EthicsGuidelines}?
    \item[] Answer: \answerYes{} 
    \item[] Justification: The paper conforms with the NeurIPS Code of Ethics.

\item {\bf Broader Impacts}
    \item[] Question: Does the paper discuss both potential positive societal impacts and negative societal impacts of the work performed?
    \item[] Answer: \answerYes{} 
    \item[] Justification: We provide an impact statement in Sec.~\ref{sec:impact}. Effectively, as with other work on generative models, they can be used for both beneficial and harmful purposes. Our work, being focused on the mechanics of these models, does not introduce new risks, nor does it mitigate existing ones.  

\item {\bf Safeguards}
    \item[] Question: Does the paper describe safeguards that have been put in place for responsible release of data or models that have a high risk for misuse (e.g., pretrained language models, image generators, or scraped datasets)?
    \item[] Answer: \answerNA{} 
    \item[] Justification: We believe that our work is not expected to have any more potential risks than other work in this field. We have discussed some of these considerations in Sec.~\ref{sec:impact}.

\item {\bf Licenses for existing assets}
    \item[] Question: Are the creators or original owners of assets (e.g., code, data, models), used in the paper, properly credited and are the license and terms of use explicitly mentioned and properly respected?
    \item[] Answer: \answerYes{} 
    \item[] Justification: For the datasets we use, we cite the original papers and describe the license information in Sec.~\ref{sec:exp-details}.

\item {\bf New Assets}
    \item[] Question: Are new assets introduced in the paper well documented and is the documentation provided alongside the assets?
    \item[] Answer: \answerYes{} 
    \item[] Justification: We provide necessary information to reproduce our new models in Sec.~\ref{sec:exp-details}.

\item {\bf Crowdsourcing and Research with Human Subjects}
    \item[] Question: For crowdsourcing experiments and research with human subjects, does the paper include the full text of instructions given to participants and screenshots, if applicable, as well as details about compensation (if any)? 
    \item[] Answer: \answerNA{} 
    \item[] Justification: The paper does not involve crowdsourcing or research with human subjects.

\item {\bf Institutional Review Board (IRB) Approvals or Equivalent for Research with Human Subjects}
    \item[] Question: Does the paper describe potential risks incurred by study participants, whether such risks were disclosed to the subjects, and whether Institutional Review Board (IRB) approvals (or an equivalent approval/review based on the requirements of your country or institution) were obtained?
    \item[] Answer: \answerNA{} 
    \item[] Justification: The paper does not involve crowdsourcing nor research with human subjects.

\end{enumerate}

\end{document}